\documentclass[10pt,twocolumn,letterpaper]{article}
\usepackage{cvpr}

\def\cvprPaperID{10824} 
\def\confName{CVPR}
\def\confYear{2022}

\usepackage{comment}

\usepackage{amsmath,amsfonts,bm}

\def\eqref#1{equation~\ref{#1}}

\def\1{\bm{1}}

\DeclareMathAlphabet{\mathsfit}{\encodingdefault}{\sfdefault}{m}{sl}
\SetMathAlphabet{\mathsfit}{bold}{\encodingdefault}{\sfdefault}{bx}{n}

\usepackage[utf8]{inputenc}
\usepackage[T1]{fontenc}
\usepackage{url} 
\usepackage{booktabs}
\usepackage{amsfonts}
\usepackage{nicefrac}
\usepackage{microtype}
\usepackage{graphicx}
\usepackage{amsmath}
\usepackage{enumitem}
\usepackage[symbol]{footmisc}
\usepackage{nicefrac}
\usepackage{siunitx}
\usepackage{multirow}
\usepackage{makecell}
\usepackage[dvipsnames,table]{xcolor}
\usepackage{etoolbox}
\robustify\cellcolor

\usepackage[pagebackref,breaklinks,colorlinks]{hyperref}

\newcommand{\set}[1]{\mathcal{#1}}

\newtheorem{defn}{Definition}

\newcommand{\bgl}{\cellcolor[HTML]{DDDDDD}}
\newcommand{\bgd}{\cellcolor[HTML]{BBBBBB}}

\renewcommand{\thefootnote}{\fnsymbol{footnote}}

\usepackage[capitalize]{cleveref}
\crefname{section}{Sec.}{Secs.}
\Crefname{section}{Section}{Sections}
\Crefname{table}{Table}{Tables}
\crefname{table}{Tab.}{Tabs.}

\begin{document}

\title{Learning ABCs: Approximate Bijective Correspondence \\for isolating factors of variation with weak supervision}

\author{Kieran A. Murphy\footnotemark[1]\\
Univ. of Pennsylvania\\
{\tt\small kieranm@}\footnotemark[2]
\and
Varun Jampani\\
Google Research\\
{\tt\small varunjampani@}\footnotemark[3]
\and
Srikumar Ramalingam\\
Google Research\\
{\tt\small rsrikumar@}\footnotemark[3]
\and 
Ameesh Makadia\\
Google Research\\
{\tt\small makadia@}\footnotemark[3]
}

\maketitle
\pagestyle{plain}
\footnotetext[1]{Work done as part of the Google AI Residency program.}
\footnotetext[2]{\texttt{seas.upenn.edu}}
\footnotetext[3]{\texttt{google.com}}

\begin{abstract}
Representational learning forms the backbone of most deep learning applications, and the value of a learned representation is intimately tied to its information content regarding different factors of variation. 
Finding good representations depends on the nature of supervision and the learning algorithm. 
We propose a novel algorithm that utilizes a weak form of supervision where the data is partitioned into sets according to certain \textnormal{inactive} (common) factors of variation which are invariant across elements of each set.
Our key insight is that by seeking correspondence between elements of different sets, we learn strong representations that exclude the inactive factors of variation and isolate the \textnormal{active} factors that vary within all sets.
As a consequence of focusing on the active factors, our method can leverage a mix of set-supervised and wholly unsupervised data, which can even belong to a different domain.
We tackle the challenging problem of synthetic-to-real object pose transfer, without pose annotations on anything, by isolating pose information which generalizes to the category level and across the synthetic/real domain gap.
The method can also boost performance in supervised settings, by strengthening intermediate representations, as well as operate in practically attainable scenarios with set-supervised natural images, where quantity is limited and nuisance factors of variation are more plentiful.
Accompanying code may be found on \href{https://github.com/google-research/google-research/tree/master/isolating_factors}{github}.
\end{abstract}

\section{Introduction}
A good representation is just as much about what it excludes as what it includes, in terms of factors of variation across a dataset~\cite{tian2020info}.
Control over the information content of learned representations depends on the nature of available supervision and the algorithm used to leverage it.
For example, full supervision of desired factors of variation provides maximum flexibility for fully disentangled representations, as an interpretable mapping is straightforward to obtain between elements and the factors~\cite{bengio13pami,Higgins2018}. 
However, such supervision is often unrealistic since many common factors of variation, such as 3D pose or lighting in image data, are difficult to annotate at scale in real-world settings. 
On the other hand, unsupervised learning makes the fewest limiting assumptions about the data but does not allow control over the discovered factors~\cite{locatello19icml}. Neither extreme, fully supervised or unsupervised, is practical for many real-world tasks. 

\begin{figure*}
    \centering
    \includegraphics[width=\textwidth]{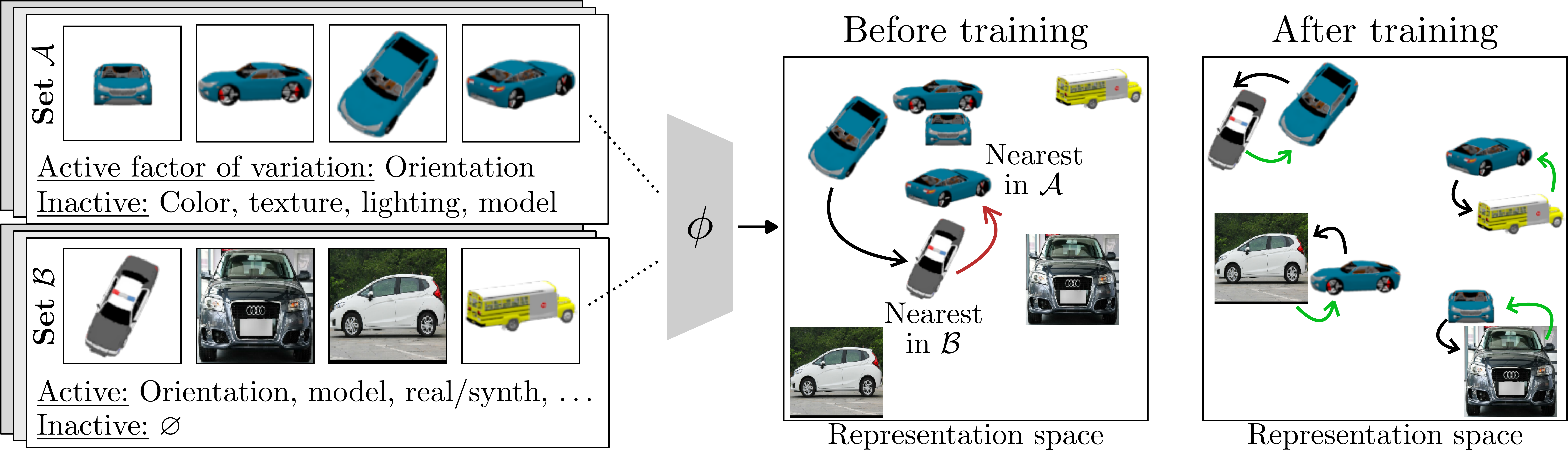}
    \caption{\small\emph{\textbf{Approximate bijective correspondence (ABC).} 
    Leveraging weak set supervision---merely groupings of data within which certain factors of variation are invariant---ABC isolates factors of variation which actively vary across sets.
    The images in set $\set{A}$ (left) actively vary by only the orientation of the rendered car.
    We claim that if one-to-one correspondence can be found between $\set{A}$ and $\set{B}$, for all possible pairs $\set{A}$ and $\set{B}$, it must leverage orientation.
    We find this to be true even when only one of the sets in each pair is set-supervised.  
    Importantly, this allows the incorporation of out-of-domain data with no supervision of any sort, such as the images of real cars in $\set{B}$. 
    By training a neural network $\phi$ with a loss that measures correspondence in representation space by the degree to which the nearest neighbor in $\set{B}$ of a point in $\set{A}$ (black arrow) is paired up with the same point in $\set{A}$ (green arrow) or a different point in $\set{A}$ (red arrow, middle), the learned representations (right) isolate the active factor of variation, orientation.
    }}
    \label{fig:teaser}
\end{figure*}

As an alternative, we consider weak supervision in the form of set membership~\cite{kulkarni15inverse,denton17nips}, used in prior works though often only informally defined.
To be specific, we assume access to subsets of training data within which some \textit{inactive} factors of variation have fixed values and the remaining \textit{active} factors freely vary for different elements of the subset.
For example, consider the images of a synthetic car in set~$\set{A}$ of Fig.~\ref{fig:teaser}.
All images in this set share common values for factors of variation relating to the specific car instance, and the only actively varying factor is the car's orientation in the image.
Set membership is the only information; there are no annotations on any factors of variation.
In many complex tasks that are beyond the scope of categorical classification, set supervision serves as a more flexible source of information for operating on factors of variation across a dataset.

Many techniques designed to utilize set supervision exploit correspondence across data that match in desired factors of variation~\cite{chen2020simple,Locatello2021}. For instance, if images of cars with the same 3D pose are grouped together (i.e. the inactive factor in each set is pose), then a straightforward training objective that maps images within groups to similar embeddings and images from different groups to dissimilar embeddings will have successfully isolated pose.
However, in this scenario and more generally, this variant of set supervision is often prohibitive to obtain: in our example it requires identifying images of different cars from exactly the same viewpoint.

A more readily available form of set supervision is where the desired factors are active in each set. 
Continuing the example, such supervision can be obtained by simply imaging each car from multiple viewpoints (as in set $\set{A}$ in Fig.~\ref{fig:teaser}). 
This does not require correspondence in viewpoints across object instances, nor any pose values attached to the images. 
However, isolating the active factors (pose in this example) from set supervision is much harder, as there is no explicit correspondence in the desired factor (i.e., no matching images with identical pose information).

In this work, our goal is to operate in this more practical set-supervised setting, but the lack of correspondence in the desired active factors makes a solution difficult. To this end, we propose a novel approach, \textit{approximate bijective correspondence} (ABC), which isolates the active factors through the process of finding correspondence between elements of different sets.
To consistently yield correspondence across sets, learned representations must ignore invariant information within a set (inactive factors) and focus on active factors common to all sets. 
A powerful consequence is the capability to incorporate sets with extraneous active factors, including wholly unsupervised and even out-of-domain data (e.g., set $\set{B}$ in Fig.~\ref{fig:teaser}), as long as one of the sets is more constrained (set $\set{A}$ in Fig.~\ref{fig:teaser}). 
In the example of Fig.~\ref{fig:teaser}, ABC-learned embeddings isolate orientation, the common active factor across every pair of sets during training.

In our approach, each element of a set is paired with a cooresponding proxy element of another set constructed with a differentiable form of nearest neighbors~\cite{Goldberger2004,Movshovitz-Attias17,Rocco2018,Snell2017,Dwibedi2019}.
The two serve as a positive pair for use in a standard contrastive (InfoNCE) loss~\cite{oord2019representation}.
We find that the same desirable properties of learned representations that optimize InfoNCE on explicitly provided positive pairs---namely, \textit{alignment}, where differences within positive pairs are ignored, and \textit{uniformity}, where maximal remaining information is retained \cite{WangIsola2020,Locatello2021}---can be utilized to guide a network to find useful correspondences on its own. 
The key strengths of ABC are the following:
\vspace{-1mm}
\begin{itemize}[leftmargin=*]
\itemsep0em 
    \item \textbf{Isolates factors inaccessible to related methods.} ABC isolates the \textit{active} factors of variation in set-supervised data, and suppresses the inactive factors.
    \item \textbf{Mixed-domain learning.} The ability to incorporate unsupervised data with extraneous factors of variation allows ABC to learn representations which bridge domain gaps with entirely unsupervised data from one domain.
    \item \textbf{Faster training.} ABC is much faster than alternative routes to isolating active factors from set-supervised data, all of which require learning the inactive factors as well.
\end{itemize}
\vspace{-1mm}
We analyze the method and its strengths through experiments on a series of image datasets including Shapes3D~\cite{shapes3d} and MNIST~\cite{mnist}.
In its fullest form, ABC addresses the challenging task of pose estimation in real images by meaningfully utilizing entirely unsupervised real images with set-supervised synthetic images, bridging the domain gap from synthetic to real.
Our experiments show that ABC presents a viable path to learning 3D pose embeddings of real images of unseen objects without having access to any pose annotations during training.
We conclude by training ABC with set-supervised real images, including one scenario matching the hypothetical example of images of cars taken from multiple viewpoints.
ABC successfully isolates active factors of variation out of the many nuisance factors of variation common to natural images, all with access to only a limited quantity of training examples.

\section{Related work}
\noindent \textbf{Isolating factors of variation.}
Recent work \cite{locatello19icml} has shown unsupervised disentanglement of latent factors to be impossible without incorporating some sort of supervision or inductive bias, spurring research into the best that can be achieved with different forms of supervision \cite{Siddharth2017,Liu2018,Shu2020Weakly,Locatello2021}.
A more realistic goal is the isolation of a subset of factors of variation, where learned representations are informative with respect to those factors and not others, with no guarantees about the structure of these factors in latent space.

\noindent \textbf{Set supervision.}
Often, data is readily grouped into sets according to certain factors of variation without requiring explicit annotation on the factors.
Generally, the methods harnessing information present in such groupings either \textbf{(i)} learn all factors and partition the representation such that one part is invariant across sets and the remaining part captures the intra-set (\textit{active}) variation~\cite{kulkarni15inverse,mathieu16neurips,cohen15iclr,sanchez20eccv,jha2018disentangling,bouchacourt18aaai}, or \textbf{(ii)} learn the factors which are invariant (\textit{inactive}) across sets~\cite{chen2020simple,tian2020info,tian2020contrastive,vowels2020nestedvae}.
The methods of \textbf{(i)} almost always employ generative models, with the exception of \cite{sanchez20eccv}, which grants it $6\times$ faster training over the VAE-based approach of \cite{jha2018disentangling}; the downside is the method of \cite{sanchez20eccv} requires seven networks and a two-stage, adversarial training process to learn first the inactive and then the active partitions of the representation.
The methods of \textbf{(ii)} generally create subsets of data via augmentation~\cite{chen2020simple,MoCo,Xiao2021WhatSN} or pretraining tasks~\cite{misra2019selfsupervised}, or leverage multiple views of the same scene \cite{TCN2017,tian2020contrastive}, where semantic information is the target of training and is taken to be invariant across sets.
By contrast, ABC directly learns \textit{active} factors of variation across sets, offering a faster and simpler alternative to methods in \textbf{(i)} and tackling problems which are currently unassailable by methods in \textbf{(ii)}.

Videos, images, and point clouds are common forms of data which naturally offer set supervision. 
Approaches to find correspondence between frames of related videos, first using a discrete form of cycle consistency~\cite{aytar2018playing} and later a differentiable form~\cite{Dwibedi2019}, helped inspire this work.
Cycle consistency has also been used to establish point correspondences in images~\cite{zhou2016learning,oron2016bestbuddies} and 3D point clouds~\cite{yang2020mapping,navaneet2020image,meshAlignmentCVPR2021}. 
In contrast to methods focusing on specific applications such as action progression in videos~\cite{Dwibedi2019,HareshCVPR2021} or robotics simulations~\cite{zhang2021learning}, we present a general approach applicable to a broad class of problems.

\paragraph{Pose estimation and domain transfer.} 
Although 3D pose estimation of objects in real images is an  actively researched topic~\cite{s2reg,mahendran17cvprw,hao6d,levinson20neurips}, supervised pose estimation is difficult to deploy in practical scenarios due to the difficulty in obtaining accurate 3D pose labels at scale, and to annotation ambiguities caused by object symmetries.  In light of the challenges posed by object symmetries, several methods attempt unsupervised learning of pose-aware embeddings rather than directly regressing absolute pose~\cite{SundermeyerAAE2018,SundermeyerAAE2020}.  In order to evaluate the learned representations, lookup into a codebook of images with known pose grants an estimate for each test image.  Others have proposed to address domain transfer where models trained on synthetic but applied on real data~\cite{ssd6d,Rad18b,wang2020self6d}; however these methods operate in constrained settings such as where the same object instance is available at both test and train time (instance-based), or exploiting depth images or 3D models for inference.
In contrast, our set-supervised method recovers pose embeddings on real images without using any pose annotations or seeing the same object instance at training time.

\section{Methods}
\label{sec:methods}
ABC uses set-supervised data, such that set membership is defined based on certain inactive factors; e.g., the data is grouped into sets such that all images in any given set have the same object class, making the object class an inactive factor. 
The basic idea of ABC is to consider all pairs of such sets (which have different values for the inactive factors of variation), and seek approximate correspondences among their elements through the learned representations.
The guiding intuition is that this can only be achieved if representations use information about the active factors of variation present in every set and exclude all other information.

\begin{figure*}[!htbp]
    \centering
    \includegraphics[width=0.95\textwidth]{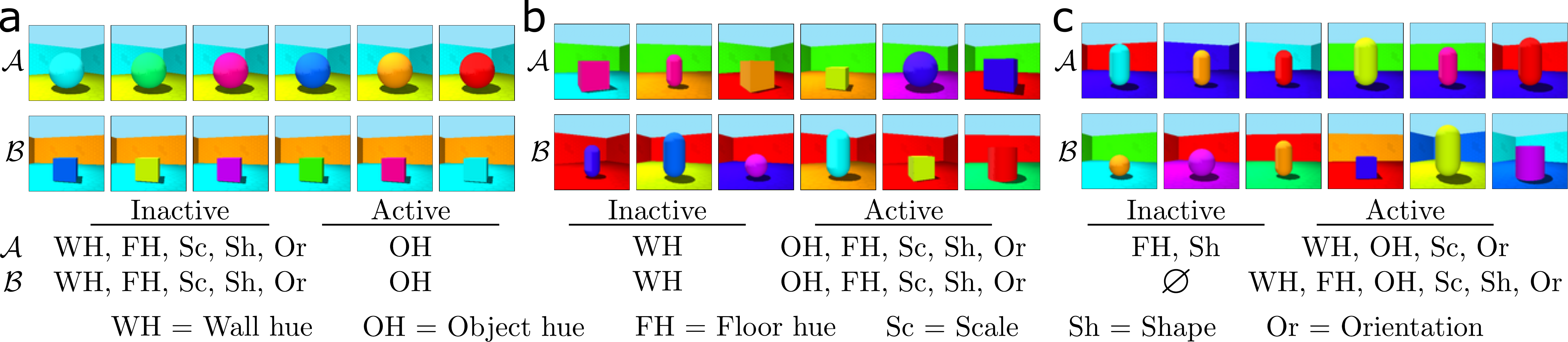}
    \caption{\small\emph{\textbf{ABC isolates active factors in a broad range of set supervision scenarios}. 
    We show an example pair of sets $\mathcal{A}$ and $\mathcal{B}$ which could arise in each of three set supervision scenarios on the Shapes3D dataset~\cite{shapes3d}.
    \textbf{(a)} In the case with five inactive factors for each set, there is only one factor to isolate and use to find correspondence: object hue.
    \textbf{(b)} The sets can be much less constrained, here defined by only a single inactive factor.  In contrast to \textbf{(a)}, all active factors may not be needed to find correspondence between every pair of sets $\mathcal{A}$ and $\mathcal{B}$.
    \textbf{(c)} One set can have extraneous active factors, or even be completely unconstrained.  In this case, correspondence is only found through active factors common to both sets, meaning floor hue and shape would not be isolated.
    In all three scenarios, ABC isolates factors which actively vary in both sets even though no correspondence is known \textnormal{a priori} between images with matching active factors.
    }}
\label{fig:factors}
\end{figure*}

To be more concrete, let us consider the pose isolation task introduced earlier. 
Assume that a latent description of each image in Fig.~\ref{fig:teaser} consists of the make and model of the car, all specifics relating to appearance, and the pose of the car in the image. 
With set-supervised data where the car instance specifics are the inactive factors within each set and the only active factor is pose (e.g., Set~$\set{A}$ in Fig.~\ref{fig:teaser}), ABC will pair elements across two sets which have similar pose.
\subsection{The ABC Algorithm}
\noindent \textbf{Setup and notation:} We follow the setup and notation from \cite{Locatello2021}, which uses a latent variable model for the theoretical modeling of self-supervised learning methods. 
Let us denote an input image as $x$ from the observation space $\set{X}$ and an associated latent code as $z$ from the representational space $\set{Z}$. 
As per the latent variable model, the observations can be generated from the latent code using an invertible function $x=f(z)$, with $z \sim p_z$. 
Without loss of generality, we assume that the latent vector $z$ can be partitioned into inactive $z_i$ and active $z_a$ components such that all elements within each set share identical $z_i$.
Let $\phi(x): \set{X} \rightarrow \mathbb{R}^E $ be the function that maps the input vector to an embedding $u$ in $E$-dimensional space. 
Our goal is to learn this function so that $u$ may be informative with respect to the active partition $z_a$ of the true underlying latent code $z$.

\noindent \textbf{Formation of pairs of sets for training:} 
We either leverage natural groupings of images or curate images into sets by controlling for certain factors of variation during mini-batch construction, where each mini-batch consists of two such sets.
For example, we show example sets with different active and inactive factors of variation curated from the Shapes3D dataset~\cite{shapes3d} in Fig.~\ref{fig:factors}. 
Values for the inactive factors are randomly sampled and held fixed for each set, with the active factors free to vary (Fig.~\ref{fig:factors}a,b).

\noindent \textbf{Approach:} 
Let the pair of sets for a particular mini-batch be given by $\set{A}=\{a_1,\dots,a_n\}$ and $\set{B}=\{b_1,\dots,b_m\}$, respectively.
Let us denote the associated embeddings as $\set{U}=\{u_1,\dots,u_n\}$ and $\set{V}=\{v_1,\dots,v_m\}$, where $u_i= \phi(a_i,w)$ and $v_i= \phi(b_i, w)$. Functionally, we parameterize $\phi$ with the same neural network (with weights $w$) for both $\set{A}$ and $\set{B}$. Let $s(u,v)$ denote a similarity metric between points in embedding space, with $s(u,v)=s(v,u)$.
To create an end-to-end differentiable loss, we use the soft nearest neighbor~\cite{Goldberger2004,Movshovitz-Attias17,Rocco2018,Snell2017,Dwibedi2019} to establish correspondence.

\begin{defn}[\emph{Soft nearest neighbor}] Given a point $u$ and a set of points $\set{V}=\{v_1,\dots,v_m\}$, the soft nearest neighbor of $u$ in the set $V$ is given by $\tilde{u} = \sum_{j=1}^m \alpha_j v_j$, where $\alpha_j = \frac{\textnormal{exp}(s(u_i,v_j)/\tau)}{\sum_{k=1}^m \textnormal{exp}(s(u_i,v_k)/\tau)}$ and $\tau$ is a temperature parameter. 
\end{defn}
We first compute the soft nearest neighbor for each $u_i \in \set{U}$ as $\tilde{u}_i = \sum_{j=1}^m \alpha_j v_j$. 
A soft bijective correspondence between the two sets is quantified through an InfoNCE loss~\cite{oord2019representation}, averaged over every element in each of the sets.
\begin{defn}[\emph{Approximate Bijective Correspondence loss}] The correspondence loss from $\set{U}$ to $\set{V}$ is given by $\mathcal{L}(\set{U}, \set{V})=-\frac{1}{n} \sum_i^n \textnormal{log} \frac{\textnormal{exp}(s(u_i,\tilde{u}_i)/\tau)}{\sum_j^n \textnormal{exp}(s(u_j,\tilde{u}_i)/\tau)}$.
The full loss is the sum, $\mathcal{L}=\mathcal{L}(\set{U}, \set{V})+\mathcal{L}(\set{V}, \set{U})$.
\end{defn}
The temperature parameter $\tau$ sets a length scale in embedding space as the natural units for the loss.
It is generally unimportant when using an unbounded similarity metric such as negative Euclidean distance (Supp.).
By contrast, a metric like cosine similarity benefits from tuning $\tau$.
\noindent \textbf{Double augmentation:}
We introduce a modification to the correspondence loss which allows suppression of factors of variation which can be augmented. 
We assume a group of transforms $H$ is known to leave desired factors of variation unchanged~\cite{Higgins2018,chenAugmentation2020,chen2020simple}.
We randomly sample two transforms $h^{(1)},h^{(2)} \in H$ per image per training step.
Let $u_i^{(1)}=\phi(h^{(1)}a_i, w)$ and similarly for $u_i^{(2)}$.
The soft nearest neighbor is found using $u_i^{(1)}$, and then the correspondence is evaluated using $u_i^{(2)}$.
The correspondence loss becomes $\mathcal{L}(\set{U}, \set{V})=-\frac{1}{n} \sum_i^n \textnormal{log} \frac{\textnormal{exp}(s(u_i^{(2)},\tilde{u}_i^{(1)})/\tau)}{\sum_j^n \textnormal{exp}(s(u_j^{(2)},\tilde{u}_i^{(1)})/\tau)}$.
The effect is to make the representations $u_i^{(1)}$ and $u_i^{(2)}$ invariant to the augmented factors of variation.

In summary, we sample pairs of sets for every mini-batch and learn an embedding network $\phi$ that produces embeddings which minimize the ABC loss through correspondence between elements in the sets.
For every element in a set, the soft nearest neighbor serves as the correspondent point in the opposite set.
The correspondence loss taken over both sets measures how close the correspondence is to being bijective.

\begin{figure*}
    \centering
    \includegraphics[width=0.93\textwidth]{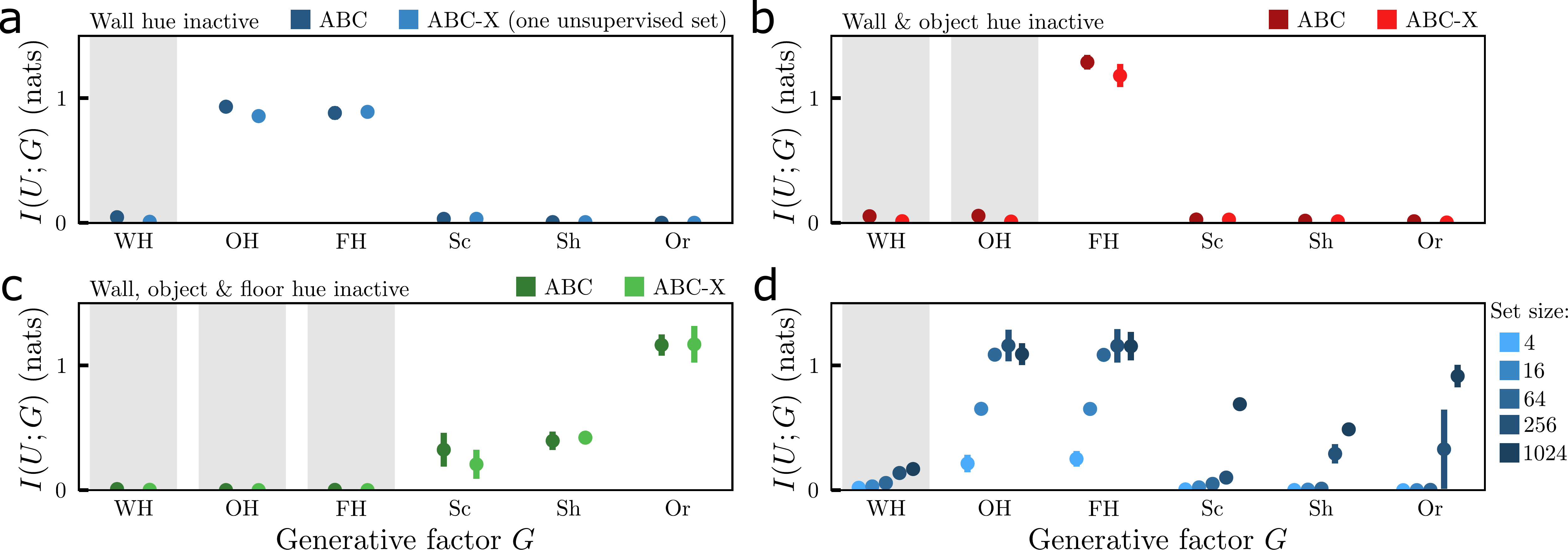}
    \caption{\small\emph{\textbf{Factor isolation even with one unsupervised set; more factors isolated with larger set sizes during training.}
    We estimate the mutual information $I(U;G)$ between the learned representations and each of the generative factors using MINE \cite{mine2018}.
    Error bars display the standard deviation across ten random seeds.
    The inactive factors during training are indicated by shading.
    \textbf{(a-c)} We find the isolation of active factors to be unchanged when training with one of the two sets unsupervised (ABC-X).
    \textbf{(d)} Increasing the set size isolates more of the active factors of variation because finding correspondence requires more discerning power.}}
    \label{fig:shapes3d_ensemble}
\end{figure*}

\subsection{Extensions}
\label{sec:extensions}
The ABC method can be extended to incorporate both fully unsupervised and supervised data.
\noindent \textbf{\textit{ABC-X} for incorporating unsupervised data:}
Only the active factors of variation common to \textit{both} sets are useful for establishing correspondence.
Information about one set's inactive factor of variation cannot help distinguish between elements of that set and therefore cannot help form correspondence with elements of another, even if the factor actively varies in the second set.
This has the powerful consequence that ABC can work just as well when one of the sets in each pair is completely unconstrained, as in Figs.~\ref{fig:teaser} and \ref{fig:factors}c.
\textit{Wholly unsupervised, and even out-of-domain data with additional active factors, can be utilized.}
We denote this version of the method ABC-Extraneous, or \textit{ABC-X}.
\noindent \textbf{\textit{ABC-S} for incorporating annotated data:}
ABC can be organically applied to an intermediate representation space in a network trained with full supervision on a particular factor of variation, by training on a weighted sum of ABC and other losses. 
If set supervision is available with the supervised factor active, ABC can condition the intermediate representation space by isolating certain factors and suppressing others, and to incorporate unsupervised data.
We denote this version of the method
ABC-Supervised, or \textit{ABC-S}.

\subsection{ABC in the context of contrastive learning} 
While both ABC and self-supervised learning (SSL) methods such as SimCLR~\cite{chen2020simple} use the InfoNCE loss on positive and negative pairs, a fundamental difference arises from how one acquires the pairs. 
In SSL a representation space is learned around explicitly provided positive pairs, obtained through augmentations known to affect certain factors while leaving others invariant. 
In ABC, a representation space is learned which also yields the positive pairs, as they are unknown \textit{a priori} and must be formed by matching nearby embeddings across sets for every evaluation of the loss.
ABC finds representations that produce good positive pairs, and does so by isolating the active factors, i.e., style, which would be inaccessible to general SSL methods. 
ABC can thus be seen as complementary to common SSL methods.

\section{Experiments}
We probe the method in four arenas.
First, we leverage complete knowledge of generative factors in the artificial Shapes3D dataset~\cite{shapes3d} in order to vary the specifics of set supervision, and precisely illustrate ABC factor isolation by measuring the information content of learned representations.
Second, we demonstrate a significant practical advantage of ABC---speed---by isolating style from class of MNIST digits~\cite{mnist}.
Third, we tackle the challenge of pose estimation on real images with no pose annotations with ABC-X, utilizing only set supervision on synthetic images.
Finally, with a limited quantity of set-supervised real images, ABC is shown to successfully isolate active factors of variation in the midst of many challenging nuisance factors.
Implementation details and extended experiments may be found in the Supp.

\subsection{Systematic evaluations on Shapes3D}
\label{sec:shapes3d}

Images from the Shapes3D dataset consist of a geometric primitive with a floor and background wall (Fig.~\ref{fig:factors}). 
There are six factors of variation in the dataset: three color factors (wall, object and floor hue) and three geometric factors (scale, shape and orientation). 
Images were grouped with certain generative factors held inactive for each of many different training scenarios in Fig.~\ref{fig:shapes3d_ensemble}; no augmentations were used.
We probed ABC-learned representations through the mutual information $I(U;G)$ between representations $U$ and known latent factors $G$,
estimated using MINE~\cite{mine2018} and averaged over ten runs each.
Deterministic networks generally preserve all information between input and output, so noise was added for a meaningful quantity $I(U+\eta;G)$, with $\eta\sim\mathcal{N}(0,\,\sigma^{2})$~\cite{saxe2019,Elad2019}.
In the case where $s(u,v)$ is negative Euclidean distance, $\tau$ serves as a natural length scale of the correspondence loss so we used $\sigma=\tau$ (Supp.).
We discuss noteworthy aspects of learned representations below.

\noindent \textit{All inactive factors are suppressed; a subset of active factors are isolated:} In Fig.~\ref{fig:shapes3d_ensemble} information with respect to all inactive factors was suppressed, and a subset of active factors---not necessarily all---were isolated.
Only when all three hue factors were inactive (Fig.~\ref{fig:shapes3d_ensemble}c) were the geometric factors present in the learned representations.
Presumably the hue factors are easier to learn and serve as shortcuts~\cite{tian2020info}, allowing the representations to ignore other factors.

\noindent \textit{Semi-supervised ABC-X is equally effective:} Correspondence is found through active factors common to both sets, which means if one set consistently has additional active factors, they will not be useful for optimizing the ABC loss.
In semi-supervised scenarios with one set-supervised set per mini-batch and the other consisting of random samples over the entire dataset (e.g., Fig.~\ref{fig:factors}c),
ABC-X performed as well as ABC with full set supervision (Fig.~\ref{fig:shapes3d_ensemble}a-c).

\noindent \textit{Increasing set size isolates more active factors:}
Intuitively, finding a one-to-one correspondence between sets with more elements requires more discerning power.
In Fig.~\ref{fig:shapes3d_ensemble}d, information in the learned representations about all active factors increased with the set size used during training.
The set size effectively serves as the number of negative samples in the InfoNCE loss, and it has been found that more negative samples benefits contrastive learning \cite{hjelm2019DIM}.

\subsection{Fast digit style isolation}
\label{sec:mnist}

Handwritten digits, such as from MNIST~\cite{mnist}, have a natural partitioning of factors of variation into digit class (e.g., 2 or 8) and style (stroke width, slant, shape, etc.). 
Our goal is to learn style information generalized across digit class, without access to style annotations or images grouped with matching style.
Images were grouped by class into sets of size 64 and embedded to $\mathbb{R}^8$; no augmentations were used.

\begin{figure}[h]
    \centering
    \includegraphics[width=0.85\columnwidth]{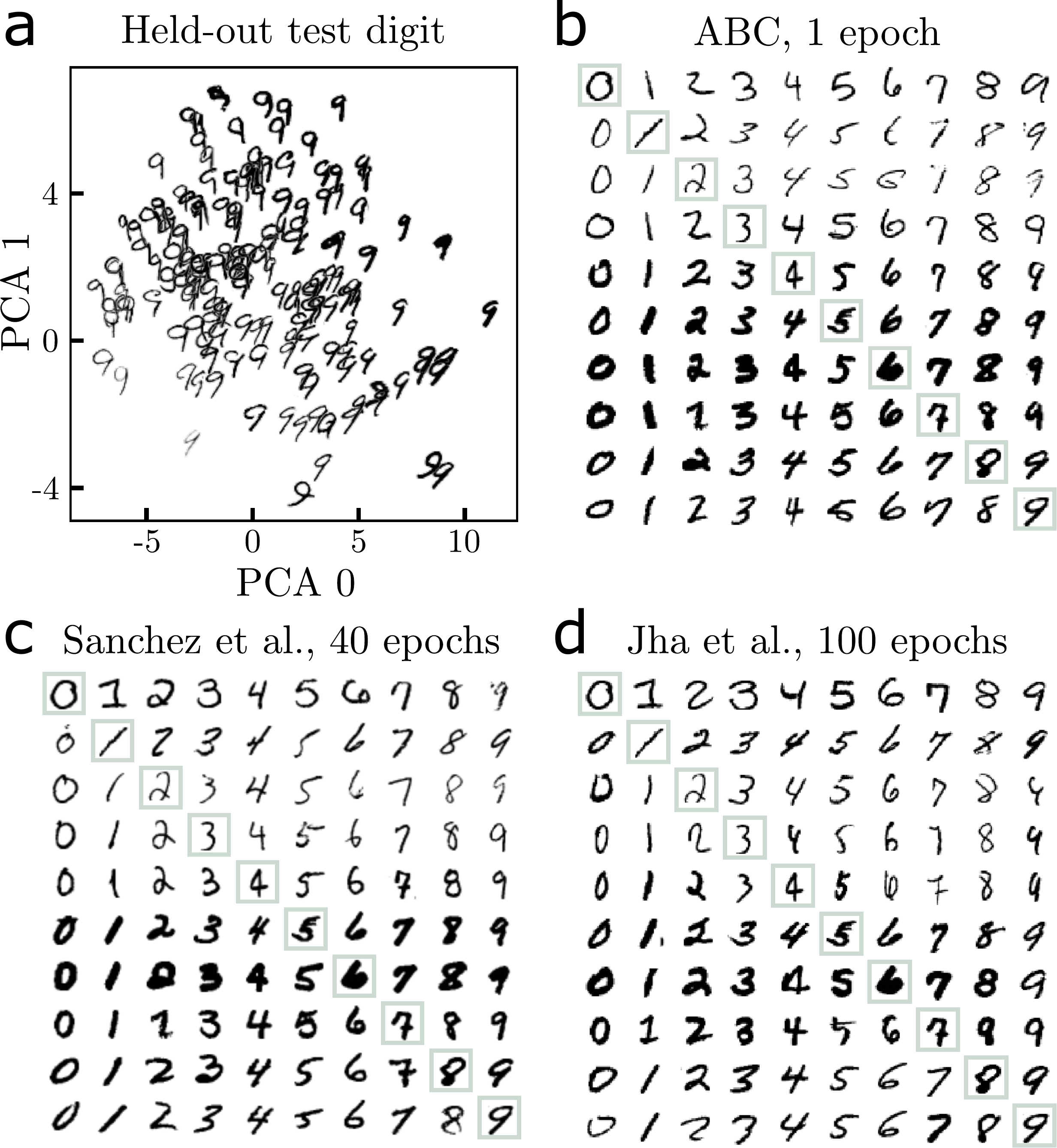}
    \caption{\emph{\textbf{Fast style isolation from MNIST digits.} 
    With digit class as the inactive factor during training, ABC isolated style. %
    \textbf{(a)} Embeddings of the digit 9---withheld during training---
    fan out by thickness and slant, active factors common to all digit classes.
    \textbf{(b)} The boxed images along the diagonal were queries for retrieval from the test set; the other images in each row were the nearest embeddings per class.  ABC isolated style information more than an order of magnitude faster than \textbf{(c)} the discriminative approach of \cite{sanchez20eccv} and \textbf{(d)} the VAE approach of~\cite{jha2018disentangling}.}
    }
    \label{fig:mnist}
\end{figure}

\begin{table*}[h]
\resizebox{\linewidth}{!}{
\centering
\small
\begin{tabular}{lccccccccc}
 & & \multicolumn{4}{c}{Cars} & \multicolumn{4}{c}{Chairs}\\
 \cmidrule(lr){3-6}\cmidrule(lr){7-10}
 & Dim ($\mathbb{R}^N$) & Med ($^\circ$) $\downarrow$ & \makecell{Acc.\ \\ @10$^\circ$ $\uparrow$} & \makecell{Acc.\ \\ @15$^\circ$ $\uparrow$} & \makecell{Acc.\ \\ @30$^\circ$ $\uparrow$} &	Med ($^\circ$) $\downarrow$ & \makecell{Acc.\ \\ @10$^\circ$ $\uparrow$} & \makecell{Acc.\ \\ @15$^\circ$ $\uparrow$}	& \makecell{Acc.\ \\ @30$^\circ$ $\uparrow$} \\
   \cmidrule(lr){2-2}\cmidrule(lr){3-6}\cmidrule(lr){7-10}
 CCVAE\cite{jha2018disentangling} & 256      & 54.9 & 0.03 & 0.07 & 0.27   & 81.5 & 0.04 & 0.07 & 0.18   \\
 ML-VAE\cite{bouchacourt18aaai} & 32  & 75.6 & 0.05 & 0.10 & 0.27  & 80.6 & 0.03 & 0.07 & 0.19    \\
 LORD\cite{gabbay2020lord} & 128  & 71.3 & 0.09 & 0.15 & 0.32  & 89.8 & 0.03 & 0.05 & 0.15    \\
 ResNet & 2048   & 85.3 & 0.07 & 0.14 & 0.28 & 80.7 & 0.04 & 0.07 & 0.19  \\
 ResNet-Intermediate & 16,384   & 15.8 & 0.30 & 0.49 & 0.64  & 47.7 & 0.08 & 0.15 & 0.37  \\
 \cmidrule(lr){1-1} Set supervision w/ TCC loss\cite{Dwibedi2019}& 64  & 23.1 & 0.14 & 0.29 & 0.59 & 58.3 & 0.09 & 0.16 & 0.40 \\
 Augmentation alone (with \cite{chen2020simple}) & 64    & 80.2 & 0.16 & 0.24 & 0.33 & 84.4 & 0.04 & 0.09 & 0.21 \\
 ABC  & 64     & \bgl{15.1} & \bgl{0.34} & \bgl{0.50} & \bgl{0.65} & \bgl{22.1} & \bgl{0.17} & \bgl{0.33} & \bgl{0.60} \\
 ABC-X  & 64 & \bgd{13.0} & \bgd{0.37} & \bgd{0.56} & \bgd{0.73} &  \bgd{16.8} & \bgd{0.27} & \bgd{0.45} & \bgd{0.74}
\end{tabular}
}
\caption{\small\emph{\textbf{Pose estimation with no pose annotations at training, set supervision on synthetic images.} Median error and accuracies (the fraction of errors better than the threshold value) on the Pascal3D+ car and chair test sets. Pose estimates were obtained through nearest neighbor lookup into a `codebook' of 1800 synthetic images with associated GT pose; reported values are the average over ten random codebooks.  The full ABC-X method---able to suppress augmentable nuisance factors of variation and to utilize unannotated real images during training---outperformed everything else, particularly in the difficult chair category.}}
\label{tab:pose}
\end{table*}

\begin{table}[t]
\centering
\small
\begin{tabular}{@{}lcccc}
 & \multicolumn{2}{c}{Cars} & \multicolumn{2}{c}{Chairs}\\
 \cmidrule(lr){2-3}\cmidrule(lr){4-5}
 & Med ($^\circ$) $\downarrow$ 	& \makecell{Acc.\ \\ @30$^\circ$ $\uparrow$} &	Med ($^\circ$) $\downarrow$ 	& \makecell{Acc.\ \\@30$^\circ$ $\uparrow$} \\
 \cmidrule(lr){2-3}\cmidrule(lr){4-5}
 Liao et al.\cite{s2reg}       & 12.3 & \bgl{0.85}  & 30.8 & 0.49    \\
 \quad + ABC-S     & \bgl{11.0} & 0.79  & \bgl{28.1} & \bgl{0.52}    \\
 \quad \makecell[l]{
 + ABC-SX
 }
 & \bgd{9.3} & \bgd{0.87} & \bgd{26.0} & \ \bgd{0.55} 
\end{tabular}
\caption{\small\emph{\textbf{Leveraging pose annotations on synthetic images, wholly unsupervised real images.} ABC-X is effective as an additional loss term when the data consists of annotated synthetic images and unannotated real images.  It provided a means to incorporate the latter which helped bridge the domain gap.
}}
\label{tab:pose_combo}
\end{table}

ABC-learned embeddings of the digit 9---withheld during training---organized according to stroke thickness and slant (Fig.~\ref{fig:mnist}a), demonstrating generalization of isolated style information across digit classes. 
In Fig.~\ref{fig:mnist}b-d we retrieved the most similar digits of each class to a set of test digits. 
Without having to learn a full description of the data, ABC yielded style-informative embeddings orders of magnitude faster than related approaches.

\subsection{Pose transfer from synthetic to real images}
\label{sec:pose}
We next utilized ABC-X for object pose estimation in real images without pose annotations at training time.
The goal was the effective isolation of pose information from set-supervised synthetic images, which generalizes to the category level and bridges the synthetic/real domain gap.
The ability of ABC-X to handle extraneous active factors of variation in one set allowed the incorporation of unsupervised real images.
This significantly extends ABC-X in Sec.~\ref{sec:shapes3d} by introducing active factors of variation which do not exist in the synthetic domain (e.g. lighting effects, occlusions).
The learned representations isolated pose, as the only factor actively varying across both sets in each training pair, while suppressing the additional domain-specific factors.

We used images of ShapeNet models~\cite{shapenet} from viewpoints randomly distributed over the upper hemisphere~\cite{Suwajanakorn2018}.
Images were grouped in sets with their source 3D model inactive (as in set $\set{A}$, Fig.~\ref{fig:teaser}). 
We gradually incorporated unsupervised real images from the Comp\-Cars~\cite{compcars} and Cars196~\cite{cars196} datasets for the car category, and 1000 images from the Pascal3D+~\cite{pascal3d} training split for chairs.
We evaluated on the test split of Pascal3D+.
All images were tight-cropped.

\begin{figure}[t]
    \centering
    \includegraphics[width=0.98\columnwidth]{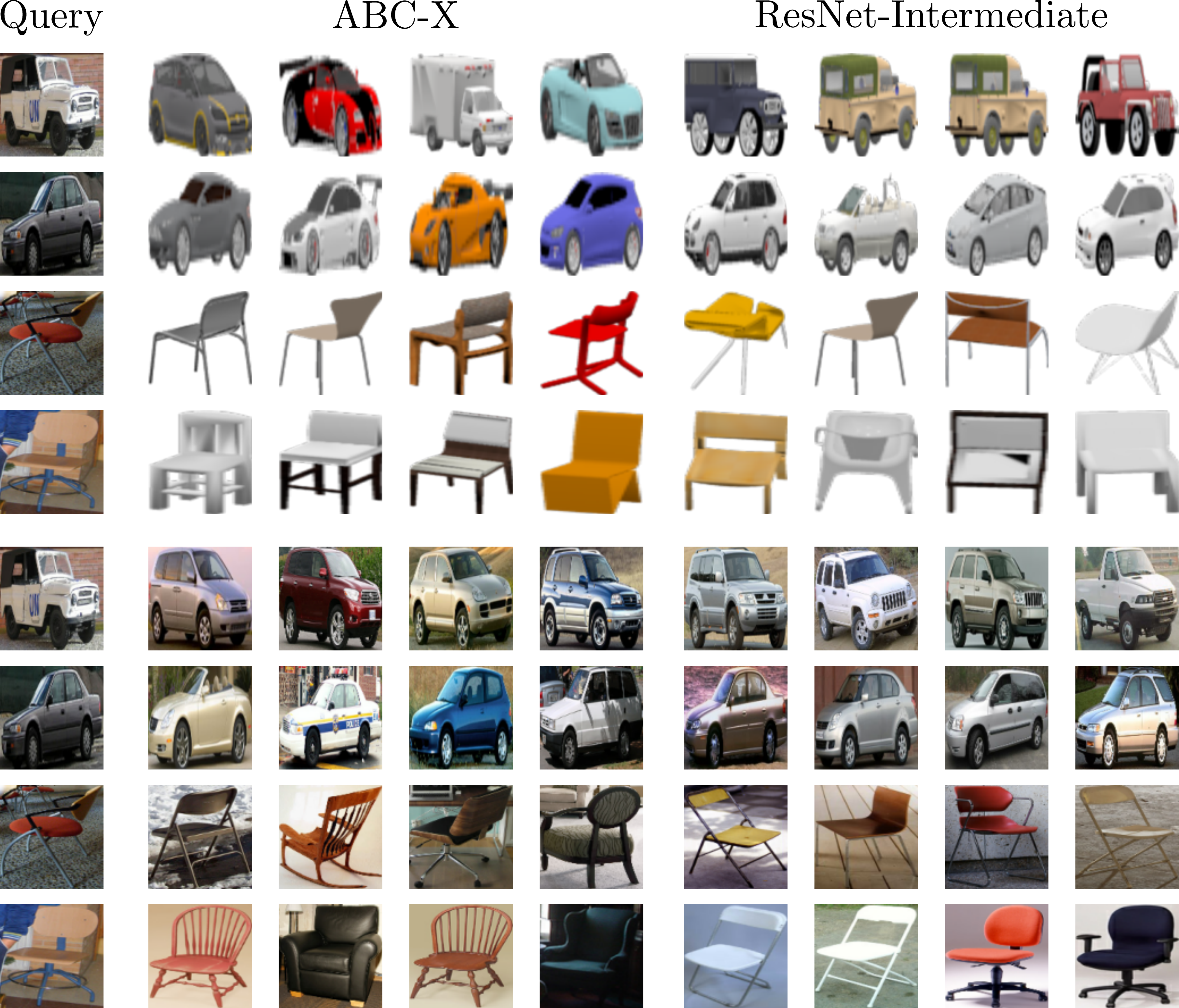}
    \caption{\small\emph{\textbf{Retrieval from ABC-X and ResNet-Intermediate.} Given a query image from the Pascal3D+ test set, we display the nearest neighbors in embedding space, from 1800 ShapeNet images and the Pascal3D+ train split. The accuracy and visual diversity of the ABC-X retrievals illustrate effective isolation of pose information generalized across the category and the synthetic/real domain gap.}}
    \label{fig:synth_retrieval}
\end{figure}

\begin{figure*}[t]
    \centering
    \includegraphics[width=\textwidth]{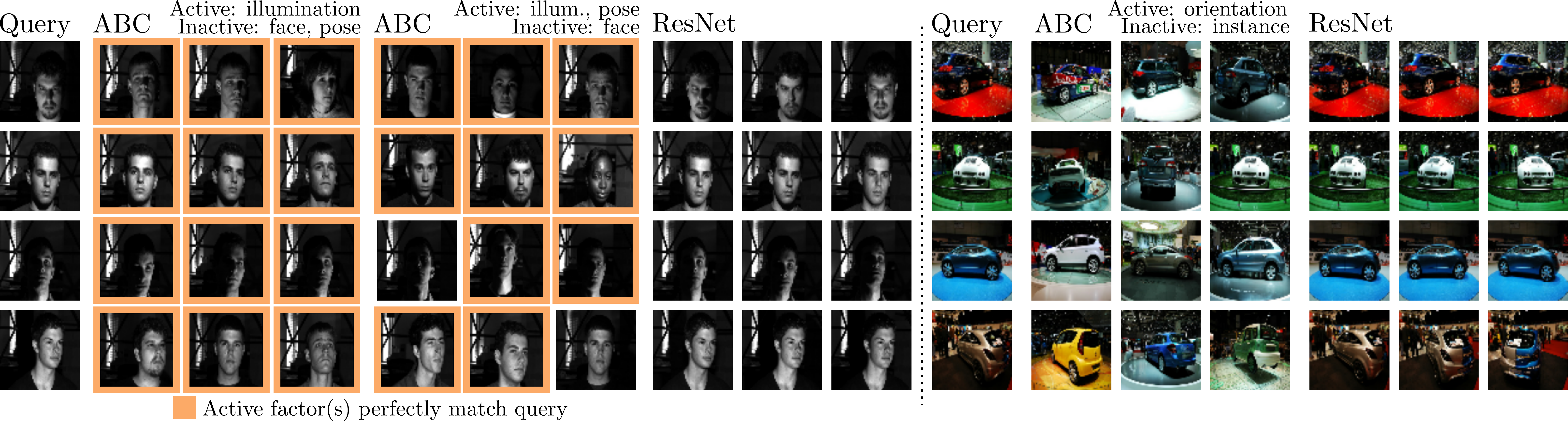}
    \caption{\small\emph{\textbf{Active factor isolation with real images only.} 
    We trained ABC on the Extended YaleB face dataset with only 28 different identities (\textbf{left}), and the EPFL Multi-View Car dataset with only 20 different car instances (\textbf{right}). 
    We compare to ResNet feature vectors, and display the nearest three neighbors (per method) to the query image out of the remainder of the dataset. Orange outlines for YaleB indicate all active factors exactly match the query image.}}
    \label{fig:real_retrieval}
\end{figure*}

The augmentation loss (Sec.~\ref{sec:extensions}) helps bridge the domain gap by removing nuisance factors of variation which could shortcut the task of finding correspondence through pose~\cite{tian2020info}.
Images were randomly augmented with cropping, recoloring, and painting the background with random crops from images of ImageNet-A \cite{hendrycks2019nae}, following augmentations used to bridge the synthetic/real domain gap in~\cite{SundermeyerAAE2018,SundermeyerAAE2020}.
Images were embedded to $\mathbb{R}^{64}$ using a few layers on top of an ImageNet-pre-trained ResNet50~\cite{resnet}.
We used cosine similarity with temperature $\tau=0.1$ (ablations in Supp.).

\subsubsection{Mixed-domain pose isolation}
\label{sec:pose1}
In the first experiment there were no pose annotations, for real nor synthetic images. 
The learned representations had no sense of absolute pose, but if pose information was successfully isolated then similar representations would have similar pose, regardless of the instance-specific details or domain of the image.
To assign a pose estimate to each image of the test set, we found the most similar synthetic image (in representation space) out of a pool of 1800, unseen at training, each with associated ground-truth pose.
We compare ABC with the VAE-based approaches of \cite{jha2018disentangling} and \cite{bouchacourt18aaai}, the latent optimization method of~\cite{gabbay2020lord}, and to feature vectors of a pre-trained ResNet (Table~\ref{tab:pose}).
We found that an intermediate output (ResNet-Intermediate), though impractical due to its high dimensionality, is a surprisingly effective baseline.

While all methods were performant when tested in the synthetic domain (Supp.), most have no means of utilizing the unsupervised real images to bridge the domain gap and consequently performed poorly when tested on real images. 
Ablative comparisons illustrate the synergy of the components of ABC-X.
Applying only the correspondence loss used in a limited setting of video alignment by~\cite{Dwibedi2019} (TCC), we found reasonable performance on the car category but a failure to isolate pose in chairs. 
Suppressing irrelevant factors from the representations via augmentation without seeking correspondence did not isolate pose for either category.
The incorporation of real images in ABC-X, ramped linearly to an average of 10\% per set $\set{B}$ by the end of training, boosted performance over ABC.
Retrieval examples (Fig.~\ref{fig:synth_retrieval}) qualitatively illustrate the generalization across instance and domain-specific factors of variation.

\subsubsection{Boosting cross-domain pose regression}
\label{sec:pose2}
We next seek to regress the pose of objects in the real domain given pose annotations in the synthetic domain only, assuming synthetic images can be grouped by instance as in Sec.~\ref{sec:pose1}. 
Starting with the spherical regression framework of~\cite{s2reg} 
we incorporated ABC-SX to condition an intermediate representation space, as described in Sec.~\ref{sec:extensions}.  
We trained on a weighted sum of a regression loss on the pose annotations for the synthetic images, and the ABC-SX loss used for Sec.~\ref{sec:pose1} (though with a subset of the augmentations).
In principle, any typical supervised pose regression network can be integrated with ABC-S.
We specifically used \cite{s2reg} as it has shown superior performance on supervised pose benchmarks, and in particular training with synthetic data (created by RenderForCNN\cite{su15iccv}) mixed with real images.

Even without real images during training, ABC-S improved performance by better conditioning the intermediate latent space (Table~\ref{tab:pose_combo}).
A further boost for both categories resulted from a small amount of real images (2\%) folded in to ABC-SX gradually over training. 
Thus ABC-SX can be advantageous in scenarios where there is more supervision available than set supervision, here serving to help bridge the real/synthetic domain gap by encouraging the suppression of factors of variation irrelevant to pose estimation.

\subsection{Set supervision in the wild}
\label{sec:realimages}
We conclude with experiments demonstrating active factor isolation with ABC from real images only.
The data is more limited in quantity and plagued by nuisance factors (such as complex backgrounds) than when the training data can incorporate a wealth of synthetic examples. 
The Extended YaleB Face dataset~\cite{lee2005yaleb} has three dominant factors of variation: face identity (of which there are only 28), face pose, and illumination orientation.
We trained the augmentation variant of ABC with illumination as the only active factor, and with both illumination and face pose as the active factors, and compare retrieval results with those from a ResNet feature vector in Fig.~\ref{fig:real_retrieval}.
Because both factors have a discrete set of possible values, the retrieval results can match the query's active factors perfectly.
We highlight perfect retrieval results in orange; ABC was remarkably successful because it can overlook person identity to find images which match in the active factor(s), something the ResNet representations failed to do.

We also trained ABC on the EPFL Multi-View Car dataset~\cite{epfl_car_turntable}, consisting of images of 20 cars on turntables with different rotation speeds, backgrounds, camera focal lengths, and rotation ranges: the hypothetical example from the Introduction.
The visual disparity of ABC-retrieval images in Fig.~\ref{fig:real_retrieval} compared to those of ResNet embeddings demonstrates the success of ABC at suppressing the many inactive factors of variation in this challenging dataset.

\section{Discussion}
The pursuit of bijective correspondence offers a powerful new foothold into factors of variation in learned representations.
ABC is significantly faster than related approaches because a full description of the data is not needed; indeed, not even all active factors of variation need be isolated.
The size of sets during training and augmentation serve as additional control over which factors of variation get isolated.
ABC is well-suited for domain transfer scenarios where an abundance of unannotated real data is accompanied by related synthetic data.
By finding its own positive pairs for use in a contrastive learning loss, ABC complements existing approaches by isolating active factors in set-supervised data. 

\noindent \textbf{Limitations:}
The task of finding correspondence does not require isolating all active factors of variation with limited set sizes, making it vulnerable to undesired ‘easy’ factors.
One should incorporate augmentations on nuisance factors if possible, and carefully analyze learned representations.

\noindent\textbf{Societal impact:}
This work is intentionally broad in its scope, and we have emphasized intuition and insight wherever possible to improve accessibility of this research.%

{\small
\bibliographystyle{ieee_fullname}
\bibliography{references}
}

\newpage

\onecolumn

\hrule
\vspace{3mm}

{\Large Supplemental Material for \textit{Learning ABCs: Approximate Bijective Correspondence for isolating factors of variation with weak supervision}}

\setcounter{section}{0}
\setcounter{page}{1}
\setcounter{figure}{0}
\setcounter{table}{0}

\renewcommand{\thepage}{S\arabic{page}}
\renewcommand{\thesection}{S\arabic{section}}
\renewcommand{\thetable}{S\arabic{table}}
\renewcommand{\thefigure}{S\arabic{figure}}
\renewcommand{\figurename}{Supplemental Material, Figure}
\renewcommand{\thefootnote}{\arabic{footnote}}

\section*{Contents}

\begin{enumerate}
\item Mutual information calculation details and supporting measurements
\item The role of length scales in isolating factors of variation
\item Synthetic-to-real pose transfer: extended results
\item Ablative studies on the pose estimation tasks
\item Augmentations used for pose estimation
\item Digit style isolation: extended results, timing measurements
\item Implementation details
\end{enumerate}

\section{Mutual information calculation details and supporting measurements} 
\label{appendix:mutual_info}
\begin{figure*}
    \centering
    \includegraphics[width=0.9\textwidth]{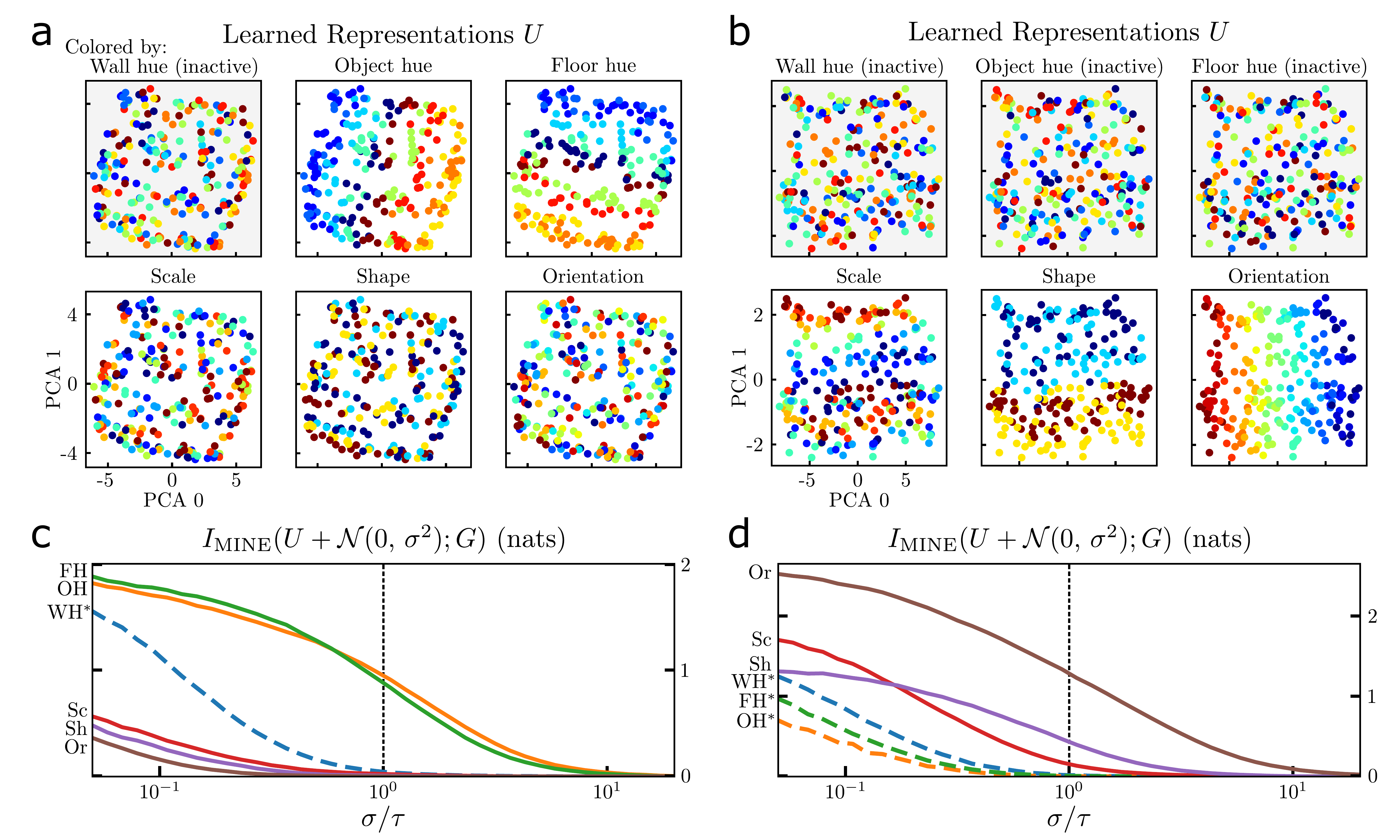}
    \caption{\small\emph{\textbf{Information content of ABC-learned representations shows active factor isolation, Shapes3D.}
    \textbf{(a)} Trained with wall hue as the only inactive factor, information about object and floor hue is visually apparent in the first two principal components ($>0.98$ of total variance) of the $\mathbb{R}^{64}$ embeddings.
    Each scatter plot displays the same 256 embeddings, colored according to each generative factor.
    \textbf{(b)} With all hue factors inactive, the representations become informative about the geometric factors.
    \textbf{(c,d)} For the networks in \textbf{(a,b)}, respectively, we estimate the mutual information $I(U;G)$ between the representations and each of the generative factors using MINE \cite{mine2018}. 
    We add Gaussian noise to the representations to probe information content over different length scales in representation space.
    When $\sigma$ equals the length scale of the loss (vertical dotted line), there is no information about inactive factors (dashed).
    }}
    \label{fig:shapes3d_scatter}
\end{figure*}

A sample of 256 Shapes3D image representations, found by ABC, are shown in Fig.~\ref{fig:shapes3d_scatter}a,b by their first two principal components.
The information about certain generative factors is qualitatively apparent, by the organization of the points by color.
Below we explain in detail how the information content was quantitatively assessed for the results of the main text.

\noindent
{\bf Calculation of mutual information.} To estimate the mutual information $I(U;G)$ for the Shapes3D experiments using MINE~\cite{mine2018}, we train a statistics network $T$.
We use a simple fully connected network whose input is the concatenation of the 64-dimensional embedding $U$ and the 1-dimensional value for the particular generative factor $G$. 
It contains three layers of 128 units each with ReLU activations, with a final one-dimensional (scalar) output.
The loss is the negated neural information measure of \cite{mine2018},
\begin{equation}
    \mathcal{L}=\text{log}(\mathbb{E}_{u\sim P(U),g\sim P(G)}[\text{exp}(T(u,g))]) - \mathbb{E}_{u,g \sim P(U,G)}[T(u,g)]
\end{equation}

At a high level, the network exploits the difference between the joint distribution $P(U,G)$, where the embedding is properly matched with its correct generative factor, and the product of marginals $P(U)P(G)$, which is simulated by shuffling the labels for the first term in the loss.
This difference between the joint and the marginals is the mutual information of the two variables.
We train with a learning rate of $3\times10^{-4}$ and a batch size of 256 for 20,000 steps, which we found to be sufficient for convergence.  
The estimate of the mutual information we report is the average value of the neural information measure over 256,000 samples from the dataset.
A new statistics network is trained for each of the six generative factors.

To handle the determinism of the embedding network, we add Gaussian distributed noise $\eta \sim \mathcal{N}(0, \sigma^2)$ directly to the embeddings.
We show sweeps over the noise scale in Figure~\ref{fig:shapes3d_scatter}c,d, where we repeat the calculation for 40 logarithmically spaced values of $\sigma$ to show the effect of this added noise on the mutual information values.
Predictably, the mutual information with respect to all factors is removed with increasing noise.
We probe the information content relevant to the InfoNCE loss when the noise and the temperature parameter $\tau$ are equal, at which point all information about inactive factors is negligible.

\begin{figure}[h]
    \centering
    \includegraphics[width=1\textwidth]{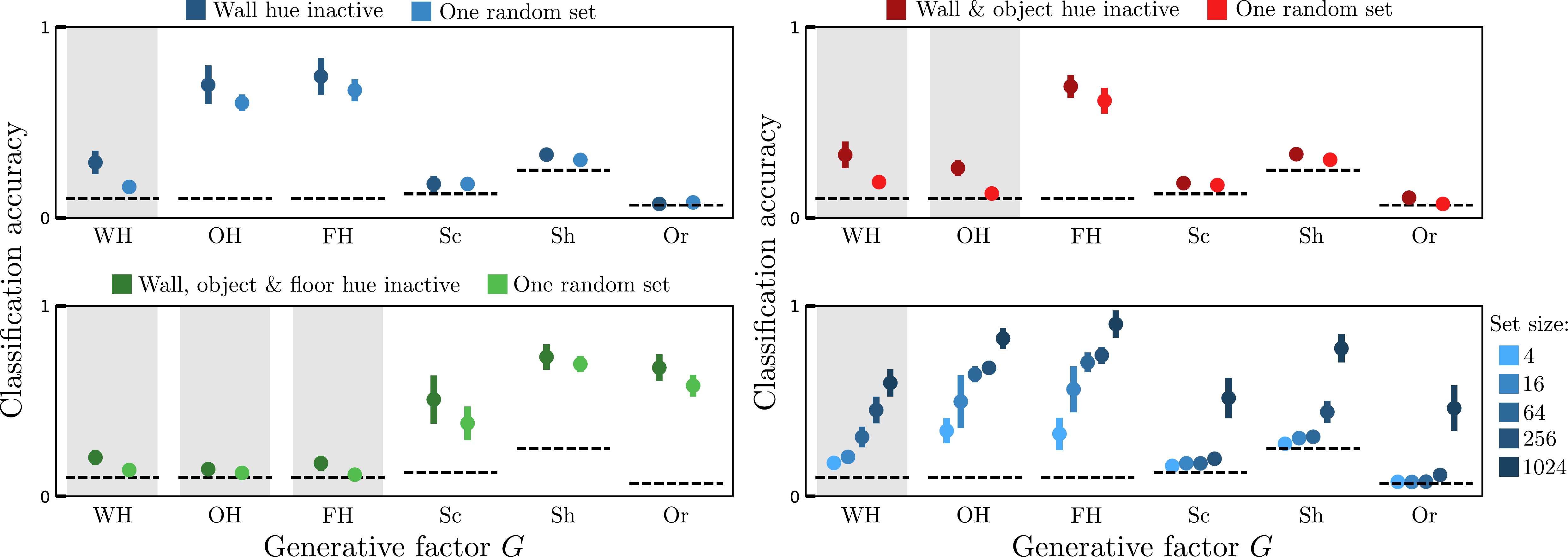}
    \caption{\emph{\textbf{Corroborating $I_\text{MINE}$ with classification task.} As a proxy for the mutual information, we use the test set classification accuracy of networks trained to predict the six generative factors, one network per factor.  As in Figure 3 of the main text, the shaded columns indicate which of the generative factors were inactive while training ABC.  Gaussian-distributed random noise with spread $\sigma$ corresponding to the length scale set by $\tau$ in the ABC loss was added to the embeddings to remove information on length scales less than the characteristic length scale of the ABC loss. The dashed lines show the classification accuracy that would result from random guessing.}}
    \label{fig:shapes_classification}
\end{figure}

\noindent
{\bf Mutual information versus classification accuracy.} To corroborate the Shapes3D mutual information measurements of Section 4.1, we use the common approach of training a simple classifier which takes the learned representations as input and tries to predict the generative factors (Figure~\ref{fig:shapes_classification}).
We train a different classifier for each generative factor, and use an architecture of 3 fully connected layers with 32 units each, ReLU activation.
As with the measurements of mutual information, there is the issue of evaluating a deterministic network which in general preserves all information~\cite{Elad2019}.
By adding Gaussian noise with magnitude $\sigma=\sqrt{\tau}$, the classification task reproduces the qualitative behavior of Figure 3.
Namely, when one or two hue factors are inactive, information about the remaining hue factor(s) is enhanced and information about the inactive factor(s) is suppressed.
When all three hue factors are inactive, then and only then is information about the three geometric factors enhanced.
There is no substantial difference in the semi-supervised setting, where one set of each mini-batch has no inactive factors.

\section{The role of length scales in isolating factors of variation}
\label{appendix:temperature}
\begin{figure}[h]
    \centering
    \includegraphics[width=0.65\textwidth]{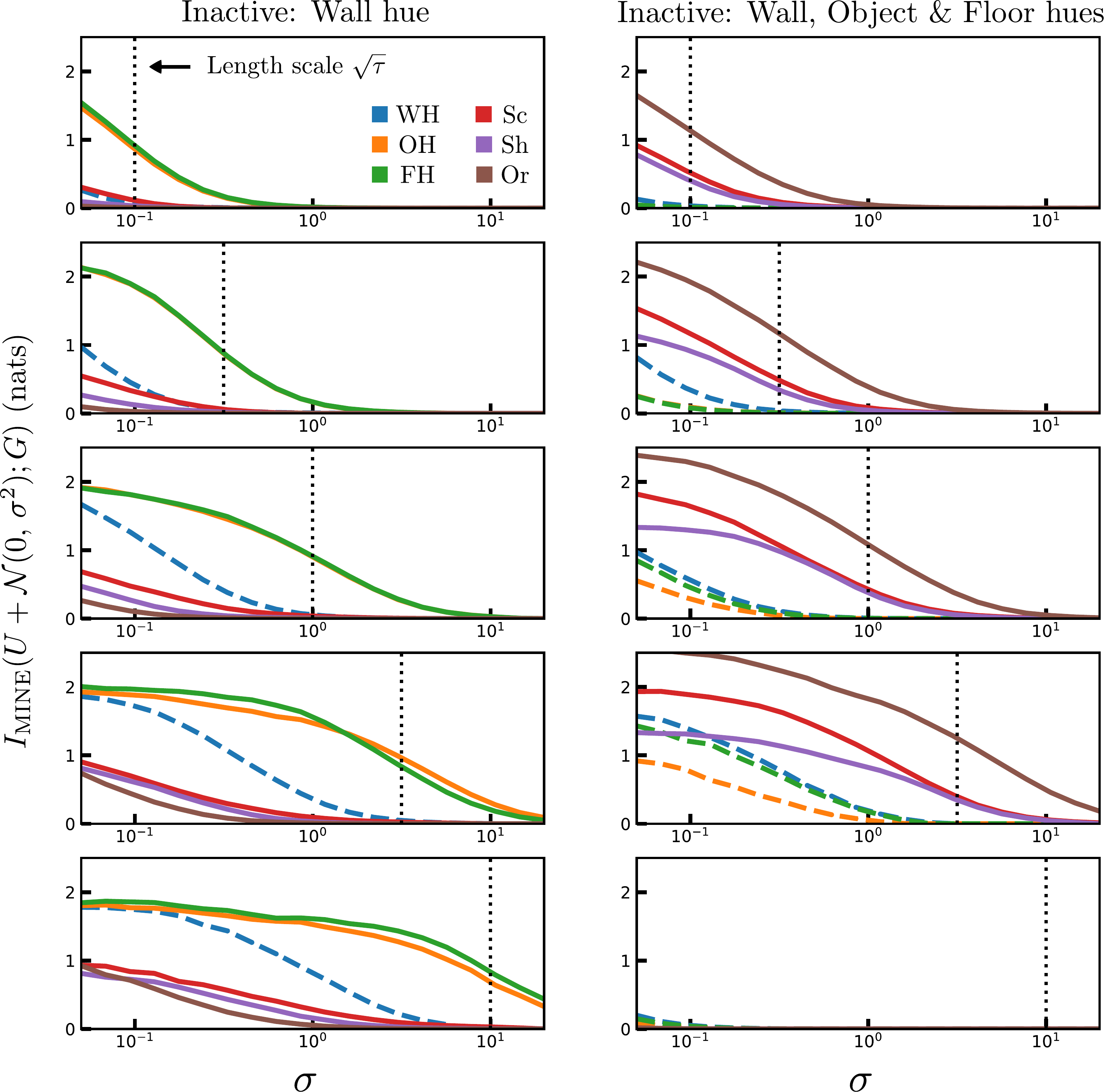}
    \caption{\emph{\textbf{Temperature sets the length scale of the cutoff between active and inactive factors.} We train with negative squared Euclidean distance between embeddings as the similarity measure, which makes $\sqrt{\tau}$ a natural length scale for embedding space.
    By varying the temperature used during training (varying vertically across the five rows), we mark the length scale $\sqrt{\tau}$ with a dotted vertical line in each subplot.
    Predictably, the magnitude of the noise $\sigma$ at which information about inactive factors is removed scales with $\sqrt{\tau}$.
    Had negative Euclidean distance been used instead, we would expect the scaling to follow $\tau$.
    The bottom right subplot shows one of the limits of varying the temperature of the ABC loss: when it is too large compared to the spread of the initialized embeddings, training is often unsuccessful.}}
    \label{fig:shapes_temp}
\end{figure}

\noindent
The ABC loss operates over a characteristic scale in embedding space, set by the temperature parameter $\tau$ which plays a role in both the soft nearest neighbor calculation and the InfoNCE loss.
When using a similarity measure derived from Euclidean distance, this characteristic scale may be interpreted as a length scale.
Two embeddings which are separated by less than this length scale effectively have a separation of zero in the eyes of the loss, and there is no incentive to further collapse them.
To be specific, when using negative L2 (Euclidean) distance as the similarity metric, the temperature $\tau$ is the characteristic length scale.
When using L2 squared distance, as in the MNIST and Shapes3D experiments, the square root of the temperature is the characteristic length scale.
With cosine similarity, as in the pose estimation experiments of Section 4.3 of the main text, temperature sets a characteristic angular difference between embeddings.

For downstream tasks, including lookup using the embeddings, this length scale is generally irrelevant. 
However, measuring the mutual information requires the addition of noise with a particular scale, and the freedom in choosing this parameter begs the question of a relevant scale in embedding space.
As a fortunate consequence, it allows a precise definition of the factor isolation that results from ABC.
We show in Figure~\ref{fig:shapes_temp} several Shapes3D experiments where the temperature $\tau$ during training took different values.
The mutual information is measured as in Figure~\ref{fig:shapes3d_scatter}c,d with a sweep over the magnitude of the added noise.

The vertical dashed line in each run shows the characteristic length scale, $\sqrt{\tau}$, and it is clear to see information about the inactive factor(s) (indicated by dashed lines) decaying to zero below the length scale.
The predicted behavior, of object and floor hue being isolated when wall hue is inactive, and of the geometric factors being isolated when all three hue factors are inactive, happens in nearly all the runs.
The length scales of everything, as measured by the magnitude $\sigma$ of the noise where the information decays, expand with increased temperature.

There is a limit to this behavior, however, which is shown in the bottom right subplot.
When the temperature is too large compared to the initial separations of the embeddings, there is too little gradient information for even the Adam optimizer to leverage, and training is unsuccessful.

\noindent
{\bf Summary.} ABC's isolation of factors has a precise meaning in representation space: 
Information about inactive factors is confined to scales less than the characteristic scale set by the temperature during training, and the isolated active factors inform the structure of embedding space over larger scales.
We demonstrate this by removing information over different scales in representation space through additive noise and mutual information measurements.

\section{Synthetic-to-real pose transfer: extended results}
\label{appendix:pose_continued}

We include in-domain pose estimation results for the car category in Table~\ref{tab:pose_synth}.
While no baseline method for the pose transfer task was particularly successful (Table 1 of the main text), all are performant when tested in the synthetic domain.
The difficulty arises from the large domain gap between synthetic and real images, and not from the ability to extract pose information from the images of the training set.

\begin{table*}[h]
\centering
\small
\begin{tabular}{lcc}
& Med ($^\circ$) $\downarrow$ & \makecell{Acc.\ \\ @30$^\circ$ $\uparrow$} \\
   \cmidrule(lr){2-3}
 CCVAE\cite{jha2018disentangling} & 12.7 & 0.67 \\
 ML-VAE\cite{bouchacourt18aaai} & \bgl{9.3} & \bgd{0.84}  \\
 LORD\cite{gabbay2020lord} & 9.9 & 0.69 \\
 ABC & \bgd{5.8} & \bgl{0.83} \\
\end{tabular}
\caption{\small\emph{\textbf{Pose estimation in synthetic domain: cars.} All baselines perform well when testing on unseen instances from the synthetic domain, highlighting the difficulty of surmounting the domain gap for the pose transfer results of Table 1 of the main text.}}
\label{tab:pose_synth}
\end{table*}

\section{Ablative studies on the pose estimation tasks}
\label{appendix:ablation}
\begin{figure*}
    \centering
    \includegraphics[width=0.9\textwidth]{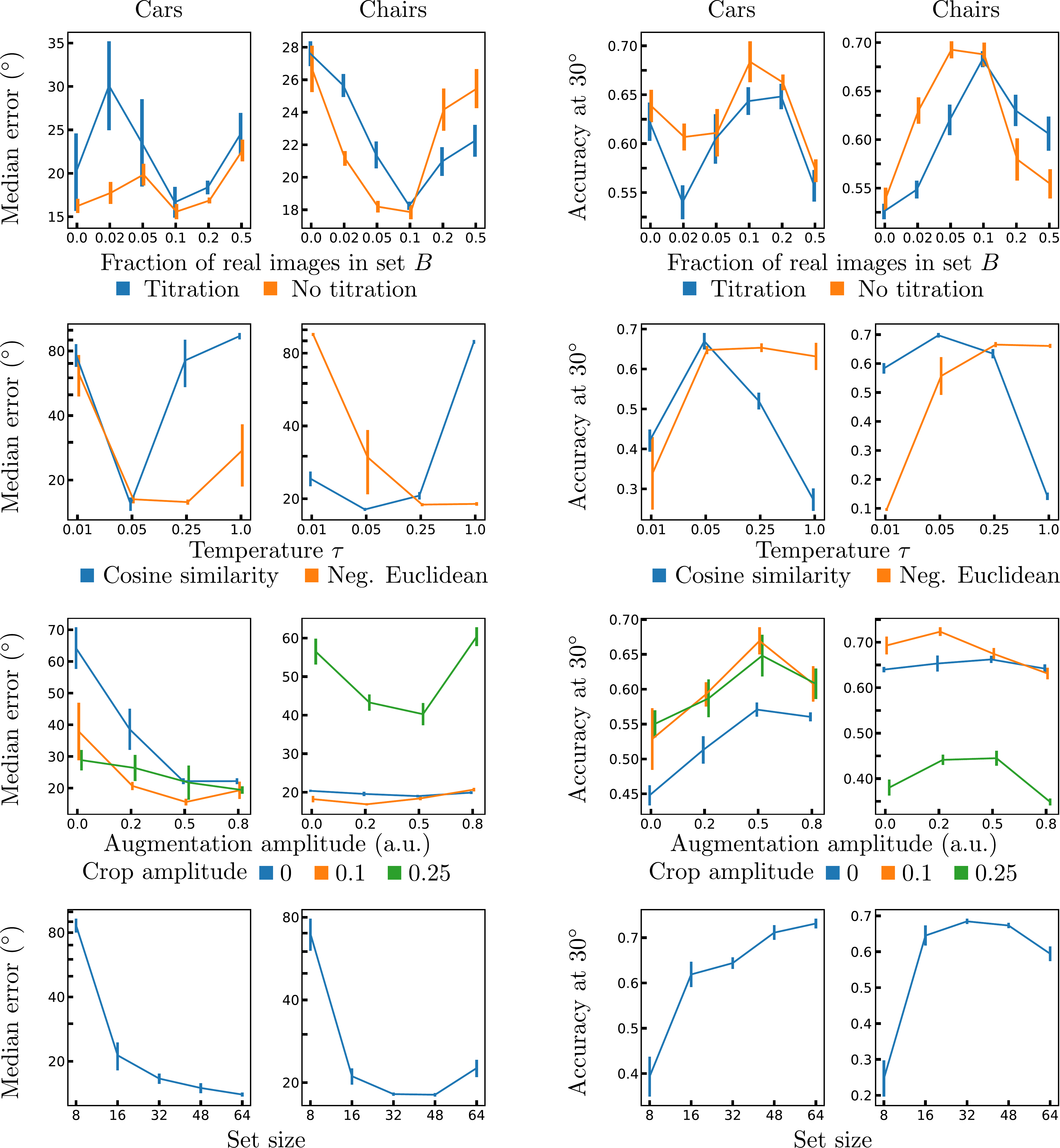}
    \caption{\emph{\textbf{Ablative studies on Pascal3D+ pose lookup with ABC embeddings.} Error bars are the standard error of the mean over 8 random seeds for each configuration.  We show results on the Pascal3D+ test split for the car and chair categories.  For each row, the training configuration is the same as described in Section~\ref{appendix:implementation} with only the listed aspect of training being changed.  In the first row, no titration means to the fraction of real images in set $B$ are present from the beginning of training. The augmentation amplitude in the third row controls the coloring changes discussed in Section~\ref{appendix:augmentation}. The crop amplitude is another form of augmentation, though we separate it for clarity.  It controls the random translation of the bounding box, as a fraction of the dimensions of the bounding box.}}
    \label{fig:ablation_lookup}
\end{figure*}

\begin{figure*}
    \centering
    \includegraphics[width=0.9\textwidth]{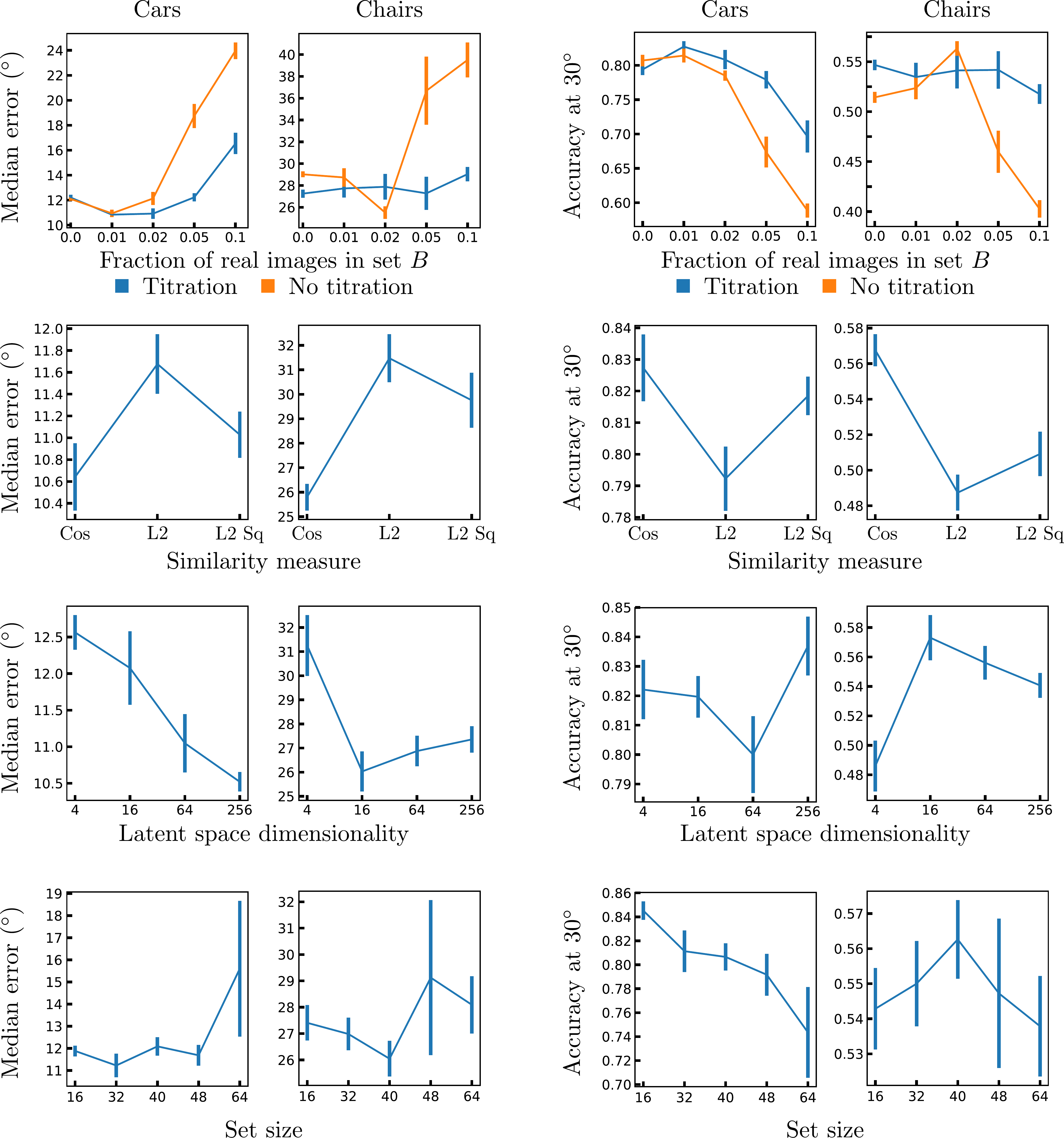}
    \caption{\emph{\textbf{Ablative studies on Pascal3D+ with spherical regression + ABC network.} Error bars are the standard error of the mean over 10 random seeds for each configuration, with less than 1\% of the runs discarded for lack of convergence.  We show results on the Pascal3D+ test split for the car and chair categories.  For each row, the training configuration is the same as described in Appendix~\ref{appendix:implementation} with only the listed aspect of training being changed.  In the first row, no titration means to the fraction of real images in set $B$ are present from the beginning of training.  The three similarity measures in the second row are cosine similarity, L2 (Euclidean) distance, and squared L2 distance.}}
    \label{fig:ablation_reg}
\end{figure*}

In Figures~\ref{fig:ablation_lookup} and~\ref{fig:ablation_reg} we show ablative studies on the pose estimation experiments of Section 4.3 of the main text, for training with the ABC loss and no pose annotations (Table 1) and the experiment where the ABC loss combined with the spherical regression method of \cite{s2reg}, utilizing pose annotations on the synthetic images (Table 2).

On both tasks, there is an optimal proportion of real images, though it is much lower for regression.
Gradual titration of real images into the unconstrained set $\set{B}$ was neutral or negative for the lookup task (Figure~\ref{fig:ablation_lookup}, top row) and generally positive for the regression task (Figure~\ref{fig:ablation_reg}, top row).
Cosine similarity outperforms negative Euclidean distance, and we show the dependence on temperature $\tau$ in the second row of Figure~\ref{fig:ablation_lookup}.

The car and chair categories present different challenges for pose estimation -- e.g. an approximate front-back symmetry for cars, greater class diversity for chairs, outdoor versus indoor settings for cars versus chairs, etc.
Several of the ablated factors cause differing effects on the performance for the two categories.

For instance, there is an apparent difference between the two categories in the dependence on the augmentation scheme, shown in the third row of Figure~\ref{fig:ablation_lookup}.
Randomly translating the bounding box by 0.1 of its height and width helps both categories, but more than that and the chair performance greatly suffers.

Another difference between the categories is seen in the final row of Figure~\ref{fig:ablation_lookup}, where increasing the set size during training only helps pose estimation on cars.
For the largest set size, however, chair pose estimation begins to suffer.
We presume the pressure to isolate more active factors of variation from increased set size can actually be harmful to the pose estimation task if unrelated factors confound the pose estimation during lookup.
Set size similarly shows mixed effects for the regression task, shown in the final row of Figure~\ref{fig:ablation_reg}.

\newpage
\section{Augmentations used for pose estimation}
\label{appendix:augmentation}
\begin{figure}[h]
    \centering
    \includegraphics[width=0.55\textwidth]{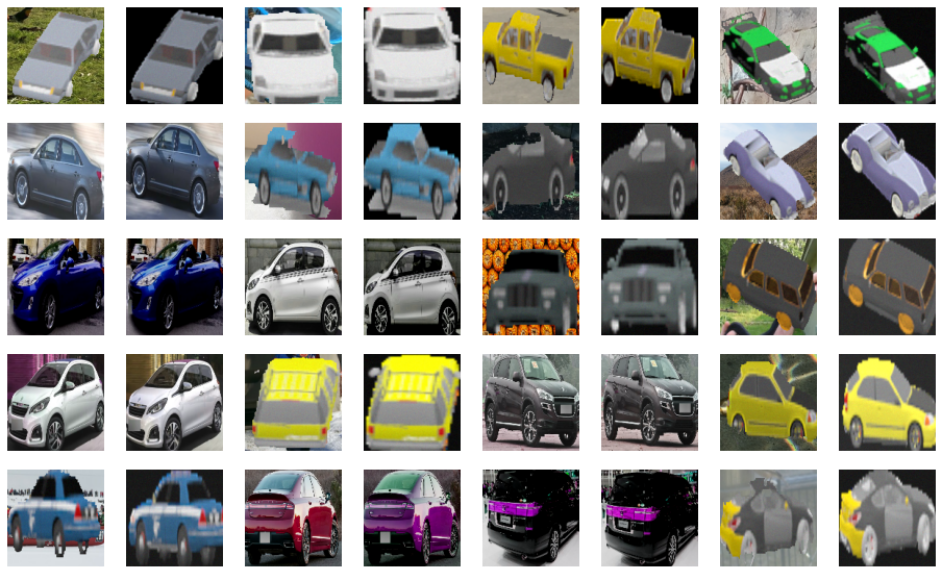}
    \caption{\emph{\textbf{Augmentations used in the pose estimation experiments.} We show sample augmentations applied to both real and synthetic cars.  These include adjusting brightness and hue, adding normally distributed noise to each pixel, random translations of the crop (bounding box), and replacing the background of synthetic images with random crops from real images.}}
    \label{fig:augmentation}
\end{figure}

For each real and synthetic image in the pose estimation tasks of Section 4.3 of the main text, we augment twice and train with the double augmentation version of the ABC loss (described in Section 3.1.1), in order to suppress additional nuisance factors from the learned representations.
We show in Figure~\ref{fig:augmentation} sample augmentation of real and synthetic car images, which include random translations of the bounding box, brightness adjustment, the addition of salt and pepper noise to each pixel, the addition of a scaled, Sobel-filtered version of the image, and hue adjustment for the real images.
We also paint the background of the synthetic images with random crops from ImageNet-A~\cite{hendrycks2019nae}.

\newpage
\section{Digit style isolation: extended results, timing measurements}
\label{appendix:mnist_continued}
\begin{figure}[h]
    \centering
    \includegraphics[width=1\textwidth]{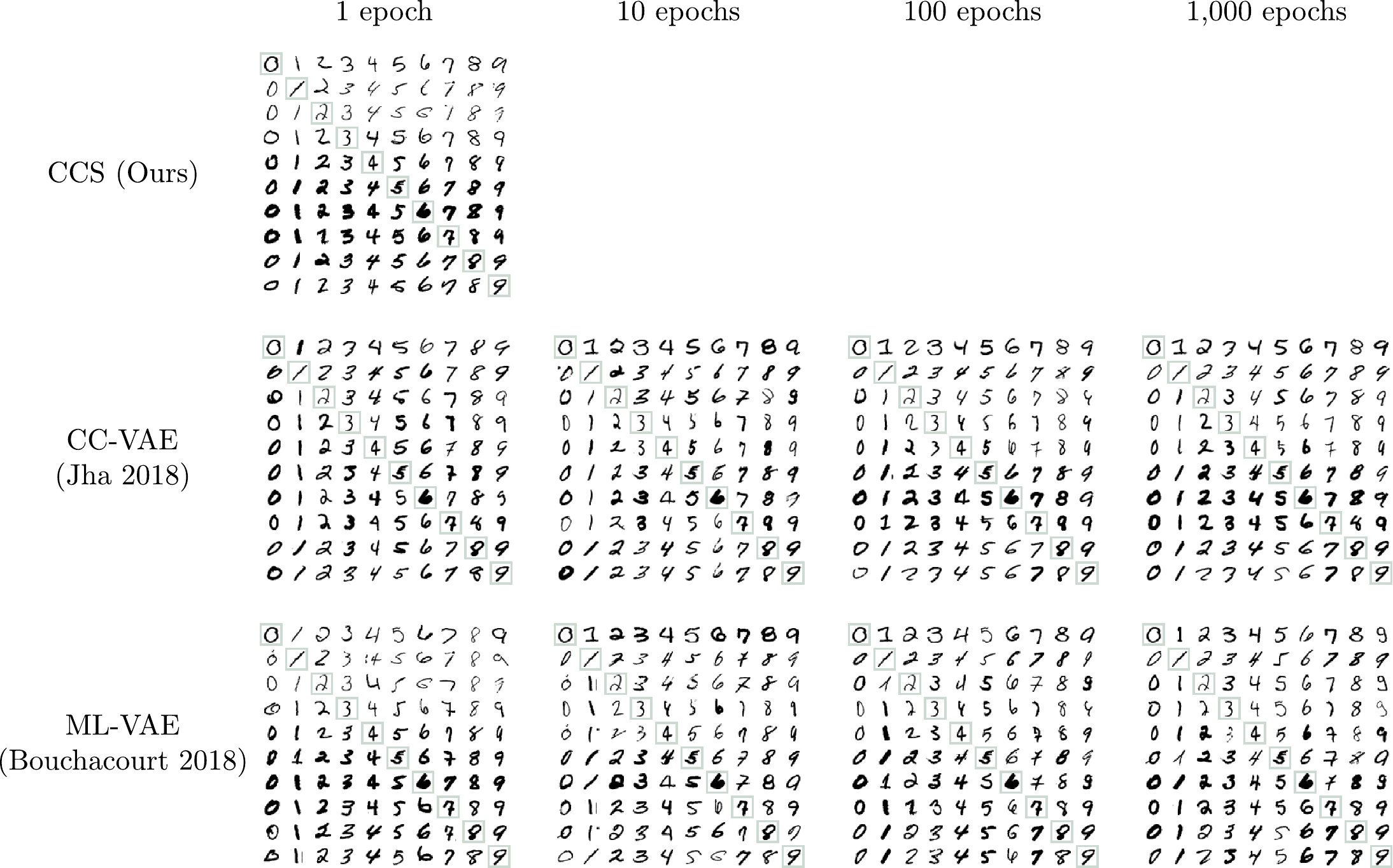}
    \caption{\emph{\textbf{Retrieval results over the course of training, comparison.} We compare retrieval on the test set of MNIST at various stages of training ABC and the two VAE-based approaches mentioned in the main text.  As in Figure 4 of the main text, the query images are the boxed images along the diagonal, and each row is the nearest representative for each class in embedding space.  Also as before, in all cases the digit 9 was withheld during training.}}
    \label{fig:mnist_extended}
\end{figure}

In Figure~\ref{fig:mnist_extended} we compare digit style isolation on MNIST using the output of ABC and the style part of the latent representations yielded by the VAE-based approaches of~\cite{jha2018disentangling} and~\cite{bouchacourt18aaai}.  
Interestingly, ML-VAE appears to embed the digits with respect to stroke thickness and slant very similarly to ABC at the beginning of training, long before any realistic images are able to be generated, but this clear interpretability of the embeddings fades as training progresses.
There are no intermediate results to show for~\cite{sanchez20eccv}, which has no style representations until the second stage of training (the last ten epochs).

\subsection{Timing calculation on MNIST}
\begin{table}[h]
{
\begin{center}
\small
\begin{tabular}{cccc}
 & ABC (ours) & Sanchez et al.~\cite{sanchez20eccv} & CC-VAE~\cite{jha2018disentangling} \\
 Seconds/epoch & \bgd{47.8} & \bgl{70.2} & 150.2 \\
\end{tabular}
\end{center}
}
\caption{\textbf{Training timing for style isolation on MNIST (Section 4.2).} These comparisons were run on an NVIDIA Tesla K80.}
\label{tab:mnist_timing}
\end{table}

We compare measurements of training time in Table~\ref{tab:mnist_timing}, all run in Tensorflow on an NVIDIA Tesla K80. 
The discriminative approaches -- ABC and \cite{sanchez20eccv} -- are far faster to train than the generative approach of \cite{jha2018disentangling}.
ABC is fastest by a wide margin due to its simplicity, requiring only one embedding network and a relatively simple loss calculation, in contrast to the seven networks and involved loss calculations required for \cite{sanchez20eccv}.

Note that by having the fastest training time per epoch, ABC further widens the gulf to the baselines, which require orders of magnitude more epochs to yield representations which isolate digit style.

\section{Implementation details}
\label{appendix:implementation}

Accompanying code may be found on \href{https://github.com/google-research/google-research/tree/master/isolating_factors}{github}.
For all experiments we use the ADAM optimizer ($\beta_1=0.9$, $\beta_2=0.999$).  Padding for convolutional layers is always `valid.'

\subsection{Shapes3D}

\begin{table*}[h]
{
\begin{center}
\small
\begin{tabular}{ccccc}
 Layer & Units & Kernel size & Activation & Stride \\
 \hline \\
 Conv2D & 32 & 3x3 & ReLU & 1 \\
 Conv2D & 32 & 3x3 & ReLU & 1 \\
 Conv2D & 64 & 3x3 & ReLU & 2 \\
 Conv2D & 64 & 3x3 & ReLU & 1 \\
 Conv2D & 128 & 3x3 & ReLU & 1 \\
 Conv2D & 128 & 3x3 & ReLU & 2 \\
 Flatten & -- & -- & -- & -- \\
 Dense & 128 & -- & ReLU & --\\
 Dense & Embedding dimension (64) & -- & Linear & --\\
\end{tabular}
\end{center}
}
\caption{\textbf{Architecture used for Shapes3D experiments (Section 4.1).} Input shape is [64, 64, 3]. }
\label{tab:shapes3d_arch}
\end{table*}

For the experiments of Figures~\ref{fig:shapes3d_scatter}\&3 we used the network architecture listed in Table~\ref{tab:shapes3d_arch}, and trained for 2000 steps with a learning rate of $3\times10^{-5}$.  We used a set size of 32 and squared L2 distance as the embedding space metric, with a temperature of 1.
To curate a set for training, we randomly sample from among the possible values for the inactive factor(s) and then filter the dataset according to it.
This takes longer when there are more inactive factors, as more of the dataset must be sieved out to acquire each stack.

\subsection{MNIST}
\begin{table}[h]
{
\begin{center}
\small
\begin{tabular}{ccccc}
 Layer & Units & Kernel size & Activation & Stride \\
 \hline \\
 Conv2D & 32 & 3x3 & ReLU & 1 \\
 Conv2D & 32 & 3x3 & ReLU & 1 \\
 Conv2D & 32 & 3x3 & ReLU & 2 \\
 Conv2D & 32 & 3x3 & ReLU & 1 \\
 Conv2D & 32 & 3x3 & ReLU & 1 \\
 Flatten & -- & -- & -- & -- \\
 Dense & 128 & -- & ReLU & --\\
 Dense & Embedding dimension (8) & -- & Linear & --\\
\end{tabular}
\end{center}
}
\caption{\textbf{Architecture used for MNIST experiments (Section 4.2).} Input shape is [28, 28, 1].  }
\label{tab:mnist_arch}
\end{table}
For the MNIST experiments we used the architecture specified in Table~\ref{tab:mnist_arch}. 
The set size was 64.  We used a learning rate of $10^{-4}$ and trained for 500 steps.  
We used squared L2 distance as the embedding space metric and a temperature of 1.
All instances of the digit 9 are held out at training time, and images of the other digits are formed into stacks before being randomly paired each training batch.
This ran in under 30 seconds on an NVIDIA Tesla V100 GPU.

\subsection{Pose estimation}
\begin{table}[h]
{
\begin{center}
\small
\begin{tabular}{ccccc}
 Layer & Units & Kernel size & Activation & Stride \\
 \hline \\
 ResNet50, up to conv4\_block6 & -- & -- & -- & -- \\
 Conv2D & 256 & 3x3 & ReLU & 1 \\
 Global Average Pooling & -- & -- & -- & -- \\
 Flatten & -- & -- & -- & -- \\
 Dense & 128 & -- & tanh & --\\
 Dense & Embedding dimension (64) & -- & Linear & --\\
\end{tabular}
\end{center}
}
\caption{\textbf{Architecture used for pose estimation experiments (Section 4.3).}  Input shape is [128, 128, 3].}
\label{tab:pose_arch}
\end{table}

For both the pose estimation lookup (Table 1) and regression (Table 2) tasks, we use the same base network to embed the images, described in Table~\ref{tab:pose_arch}.  
In contrast to the Shapes3D and MNIST experiments, we train with mini-batches consisting of 4 pairs of image sets, each of size 32.  
We use cosine similarity and a temperature of 0.1 for lookup and 0.05 for regression.
For the lookup task, the network trained for 40k steps with a learning rate that starts at $10^{-4}$ and decays by a factor of 2 every 10k steps.
The beginning of training is purely synthetic images and then ramping up linearly to 10\% real images folded into the unconstrained stack, stepping every 4k steps.

For regression, the embeddings are then fed, separately for each Euler angle, as input to a 128 unit dense layer with tanh activation, which is then split off into two dense layers with 2 and 4 units and linear activation for the angle magnitude and quadrant, respectively, as in \cite{s2reg}.
To maintain consistency between how the embeddings are processed for the ABC loss and how they are fed into the regression sub-network, the embeddings are L2-normalized to lie on the 64-dimensional unit sphere before the regression.
The angle magnitudes are passed through a spherical exponential activation function \cite{s2reg}, which is the square root of a softmax.
The magnitudes are then compared with ground truth $(|\text{sin} \phi_i|, |\text{cos} \phi_i)|$, with $i$ spanning the three Euler angles, through a cosine similarity loss.
The quadrant outputs are trained as a classification task with categorical cross entropy against the ground truth angle quadrants, defined as $(\text{sign}(\text{sin} \phi_i), \text{sign}(\text{cos} \phi_i))$.
Training proceeds for 60k steps with a learning rate that starts at $10^{-4}$ and decays by a factor of 2 every 20k steps.

To more closely match the distribution of camera pose in real images, we filter the ShapeNet renderings by elevation: 0.5 radians and 1.3 radians for the max elevation for cars and chairs, respectively (for all of the baselines as well).

\subsection{Factor isolation on real images only}
\begin{table}[h]
{
\begin{center}
\small
\begin{tabular}{ccccc}
 Layer & Units & Kernel size & Activation & Stride \\
 \hline \\
 Conv2D & 64 & 3x3 & ReLU & 1 \\
 Conv2D & 64 & 3x3 & ReLU & 2 \\
 Conv2D & 128 & 3x3 & ReLU & 1 \\
 Conv2D & 128 & 3x3 & ReLU & 2 \\
 Conv2D & 256 & 3x3 & ReLU & 1 \\
 Conv2D & 256 & 3x3 & ReLU & 2 \\
 Flatten & -- & -- & -- & -- \\
 Dense & 128 & -- & ReLU & --\\
 Dense & Embedding dimension (64) & -- & Linear & --\\
\end{tabular}
\end{center}
}
\caption{\textbf{Architecture used for the Extended YaleB Face and EPFL Multi-view Car experiments (Section 4.4).} Input shape is [64, 64, 1] for the face images and [64, 64, 3] for the car images.  }
\label{tab:mnist_arch}
\end{table}

\subsubsection{Extended YaleB Face}
Images of all 28 people were used for training.
There are 9 poses per person and 64 illumination angles.
With the illumination as the only active factor, the full set of 64 images was used for each set; with illumination and pose active, 384 images were randomly sampled for each person (out of 576) every time a set was made for training.

Each image was augmented twice with a random shift of contrast, and then a random crop (though the faces were not tight-cropped for this dataset).
We used negative L2 distance as the similarity measure, and a learning rate of $3\times10^{-4}$ for 10,000 steps (illumination active) or 50,000 steps (illumination and pose active).

\subsubsection{EPFL Multi-view Car}
We used a set size of 81 images randomly sampled from the sequence of images for a particular car.
All 20 car sequences were used for training.
Each image was augmented twice with a random shift of contrast and saturation, and then a random crop (though the cars were not tight-cropped for this dataset).
We used negative L2 distance as the similarity measure, and a learning rate of $3\times10^{-4}$ for 10,000 steps.

\subsection{Baselines}
Imagenet-pretrained ResNet:
We use the same ResNet50V2 base as for the ABC embedding network, and compare representations for each image by cosine similarity (which performed better than comparing by L2 distance).

Sanchez et al.~\cite{sanchez20eccv}:
We used the colored-MNIST architecture specifications and hyperparameters described in the Supplemental Material for the MNIST experiments of Section 4.2.
As the colored-MNIST factors of variation isolated by \cite{sanchez20eccv} are simpler in nature (color of foreground/background from specific digit, versus digit identity from style), we found better results by boosting the dimension of the exclusive representation to 64 (up from the original 8 for the color description).

CCVAE~\cite{jha2018disentangling} and ML-VAE~\cite{bouchacourt18aaai}:
We translated the publicly available pytorch code to tensorflow for training MNIST\footnote{\url{https://github.com/ananyahjha93/cycle-consistent-vae}},\footnote{\url{https://github.com/DianeBouchacourt/multi-level-vae}}.
We were unable to find code for their experiments on larger image sizes, but we followed the encoder and decoder specifications for the 64x64 RGB images in the Supplemental for \cite{jha2018disentangling}, found here\footnote{\url{https://arxiv.org/pdf/1804.10469.pdf}}, for both methods.
We optimized hyperparameters in a grid search around the published numbers, and used a group size for \cite{bouchacourt18aaai} which matched the stack size used for the ABC method.
As with \cite{sanchez20eccv}, we downsized the ShapeNet renderings and Pascal3D+ tight crops to 64x64, after attempts to scale the encoder-decoder architecture up to 128x128 were unsuccessful.

LORD~\cite{gabbay2020lord}:
We ran the publicly available tensorflow code with the parameters for the paper's Cars3D experiments\footnote{\url{https://github.com/avivga/lord}}.
Because a separate latent vector for each instance must be learned, memory became an issue when using the full training set (as used for all other methods).
We used 200 instances each for cars and chairs to closely match the 183 Cars3D instances used in their work, with images downsized and tight-cropped to 64x64 to match this works' architecture.

\end{document}